\documentclass[pdflatex,sn-nature]{sn-jnl}

\usepackage{graphicx}%
\usepackage{multirow}%
\usepackage{amsmath,amssymb,amsfonts}%
\usepackage{amsthm}%
\usepackage{mathrsfs}%
\usepackage{cite}
\usepackage[title]{appendix}%
\usepackage{xcolor}%
\usepackage{textcomp}%
\usepackage{manyfoot}%
\usepackage{natbib}
\usepackage{booktabs}%
\usepackage{algorithm}%
\usepackage{algorithmicx}%
\usepackage{algpseudocode}%
\usepackage[most]{tcolorbox} 
\usepackage{listings}%
\usepackage{makecell}  
\usepackage{booktabs}  
\usepackage{rotating}  
\usepackage{array}     
\usepackage{subcaption}
\theoremstyle{thmstyleone}%
\theoremstyle{thmstyletwo}%

\theoremstyle{thmstylethree}%

\begin{document}


\title[Article Title]{Active Evidence-Seeking and Diagnostic Reasoning in Large Language Models for Clinical Decision Support}
\author[1]{\fnm{Chen} \sur{Zhan}}\email{M325124527@sues.edu.cn}
\equalcont{These authors contributed equally to this work.}
\author*[1]{\fnm{Xihe} \sur{Qiu}}\email{qiuxihe1993@gmail.com}
\equalcont{These authors contributed equally to this work.}
\author[2]{\fnm{Xiaoyu} \sur{Tan}}\email{txywilliam1993@outlook.com}
\equalcont{These authors contributed equally to this work.}

\author[3]{\fnm{Xibing} \sur{Zhuang}}\email{19111210024@fudan.edu.cn}
\equalcont{These authors contributed equally to this work.}

\author[1]{\fnm{Gengchen} \sur{Ma}}\email{m320123320@sues.edu.cn}
\author[1]{\fnm{Yue} \sur{Zhang}}\email{m325123614@sues.edu.cn}

\author[4]{\fnm{Shuo} \sur{Li}}\email{shuo.li11@case.edu}

\author[5]{\fnm{Peifeng} \sur{Liu}}\email{lpf@sjtu.edu.cn}
\author*[6]{\fnm{Xiaoxiao} \sur{Ge}}\email{tinagxx.zs@163.com}

\author*[6]{\fnm{Liang} \sur{Liu}}\email{liu.liang1@zs-hospital.sh.cn}

\author*[7]{\fnm{Lu} \sur{Gan}}\email{gan.lu@zs-hospital.sh.cn}

\affil[1]{\orgdiv{School of Electronic and Electrical Engineering}, \orgname{Shanghai University of Engineering Science}, \orgaddress{ \city{Shanghai}, \postcode{201620}, \country{China}}}

\affil[2]{\orgdiv{Department}, \orgname{Tencent Youtu Lab}, \orgaddress{ \city{Shanghai}, \postcode{200233}, \country{China}}}

\affil[3]{\orgdiv{Department of Oncology}, \orgname{Jinshan Hospital}, \orgaddress{\street{Fudan University}, \city{Shanghai}, \postcode{201508}, \country{China}}}

\affil[4]{\orgdiv{Departments of Biomedical Engineering, and Computer and Data Science},\orgaddress{Case Western Reserve University \city{Cleveland}, \postcode{44106}, \state{Ohio}, \country{USA}}}

\affil[5]{\orgdiv{State Key Laboratory of Systems Medicine for Cancer}, \orgname{Shanghai Cancer Institute, Renji Hospital}, \orgaddress{\street{School of Medicine, Shanghai Jiao Tong University}, \city{Shanghai}, \postcode{200032}, \country{China}}}

\affil[6]{\orgdiv{Integrative Clinical Research Ward, Clinical Medicine Research Institute}, \orgname{Zhongshan Hospital}, \orgaddress{\street{Fudan University}, \city{Shanghai}, \postcode{200032}, \country{China}}}


\affil[7]{\orgdiv{Department of Medical Oncology}, \orgname{Zhongshan Hospital}, \orgaddress{\street{Fudan University}, \city{Shanghai}, \postcode{200032}, \country{China}}}


\abstract{Large language models perform well on static medical examinations, yet clinical diagnosis often requires iterative evidence gathering under uncertainty. Building on prior interactive evaluation efforts, we introduce an OSCE-inspired standardized patient simulator and a controlled, reproducible benchmark for active diagnostic inquiry. Across 468 cases and 15 models in our protocol, we observe that multi-turn evidence seeking reduces diagnostic accuracy by 12.75\% and lowers supporting-evidence quality by 24.36\% relative to full-context evaluation; error analyses associate these drops with premature diagnostic closure and inefficient questioning. Together, these results suggest that static full-context benchmarks may overestimate performance in interactive evidence-seeking settings, motivating complementary interactive assessment for safer clinical decision support.}

\keywords{Large language models, Clinical decision support, Standardized patient simulation, Interactive diagnostic reasoning}



\maketitle

\section{Introduction}\label{sec1}
In recent years, general-purpose large language models (LLMs) have demonstrated strong capabilities in medical knowledge retrieval and reasoning \citep{liao2024automatic, idan2025primer, sandmann2024systematic}. Results from static medical benchmarks such as MedQA-USMLE, PubMedQA, and MedMCQA suggest that models ranging from ChatGPT to medically specialized variants such as Med-PaLM 2 can meet or exceed passing thresholds on standardized licensing-style examinations \citep{jin2021disease,jin2019pubmedqa, pal2022medmcqa, singhal2025toward, nori2023capabilities, brin2023comparing}. These achievements have motivated growing interest in using LLMs for Clinical Decision Support Systems. However, many widely used evaluation settings remain largely passive and full-context: complete case information, including patient history, symptoms, and test results, is provided to the model simultaneously before it generates a response \citep{liu2023medical, bedi2025testing}. This setup is well-suited for assessing summary interpretation given complete information, but it does not directly test how models decide what evidence to request when information is initially sparse. As a result, it can blur the distinction between using provided facts and actively reasoning under uncertainty in iterative clinical workflows.

In practice, clinical diagnosis is a dynamic, multi-turn interactive process governed by the hypothetico-deductive model \citep{coderre2003diagnostic}. Clinicians typically do not receive all relevant data points instantaneously, especially early in a workup. Instead, they must actively formulate initial hypotheses and seek evidence through successive exchanges involving history taking, physical examinations, and laboratory testing to dynamically confirm or refute differential diagnoses \citep{elstein1978medical}. Medical education often uses the Objective Structured Clinical Examination (OSCE) to assess clinicians during such active reasoning processes \citep{harden1975assessment, khan2013objective}. By utilizing standardized patient simulations, the OSCE evaluates key competencies including active inquiry, information acquisition strategies, and the ability to synthesize evolving data \citep{barrows1993overview}. This creates a partial mismatch between real diagnostic workflows and many existing evaluation setups: static information feeds can underrepresent the role of active evidence acquisition and iterative decision-making.

Several recent studies have moved toward more clinically grounded evaluations by introducing interactive inquiry or multi-step diagnostic procedures \citep{yao2024medqa, jiang2025medagentbench, williams2024evaluating}. For example, MediQ proposed a simulated clinical dialogue framework to assess information gathering through question-answer interactions, highlighting that under incomplete initial information even large models may struggle to generate efficient and clinically meaningful follow-up questions \citep{li2024mediq, liao2024automatic}. Other efforts structure evaluation into staged diagnosis such as preliminary diagnosis, differential diagnosis, and final diagnosis to better reflect sequential decision making. Nonetheless, these approaches involve practical trade-offs, and it is not always straightforward to isolate information acquisition strategies from diagnostic reasoning performance: performance can be influenced by open-ended dialogue variability, prompting choices, and scoring criteria, which can make it difficult to determine whether errors stem from suboptimal evidence seeking or from evidence synthesis \citep{chen2020meddialog, saley2024meditod, tsoukalas2015data,von2025evaluation}. In addition, work examining conversational quality and interpretability further suggests that static question-answer benchmarks do not adequately characterize performance during realistic clinical interactions.

To better approximate authentic diagnostic processes, some initiatives have explored system designs that support cyclical inquiry and dialogue management \citep{markus2021role}. For instance, the Articulate Medical Intelligence Explorer, known as AMIE, studied an optimized conversational diagnostic system to evaluate performance across multiple dimensions including history collection, diagnostic accuracy, and communication skills \citep{tu2025towards}. Another representative work, Ask Patients with Patience, structured multi-turn inquiry into iterative dialogues driven by medical guidelines \citep{zhu2025ask, walker2024artificial}. While these studies demonstrate the value of dialogue-oriented designs, they are often optimized for specific systems, and their evaluation protocols may not directly provide a standardized, reusable benchmark. These systems often rely on open-ended conversations, which can complicate reproducibility and objective grading without human intervention \citep{zheng2023judging}. Consequently, there remains a need for benchmarks that combine controlled information release, reproducibility, and objective scoring-especially for separating evidence acquisition from evidence synthesis for systematically evaluating model capabilities in active evidence seeking and clinical reasoning within a controlled and reproducible environment that can isolate specific failure modes \citep{ke2025clinical, griot2025large}.

To complement existing static and dialogue-based evaluations, we introduce an OSCE-inspired interactive framework, ROUNDS-Bench. This framework incorporates a Standardized Patient Simulator and controls information release through a multi-turn inquiry design, providing evidence only when the model submits logical clinical requests \citep{gaber2025evaluating, luo2025large}. Compared with open-ended chat evaluations, this request–response protocol improves controllability and reproducibility. It resembles common query workflows (e.g., electronic health record retrieval), enabling measurement of questioning efficiency and diagnostic performance across turns during both the information acquisition and reasoning phases separately\citep{jiang2025medagentbench}.

Using 468 diverse cases covering six major clinical systems, we designed a dual-task evaluation structure to quantify the performance gap between full-context and active inquiry settings. Task 1, Full-Context Diagnosis, establishes the upper bound of model performance under conditions of complete information, whereas Task 2, Active Evidence Seeking, requires the model to autonomously drive a multi-turn diagnostic process starting with only the chief complaint. A series of systematic evaluations on current state-of-the-art Large Language Models, including GPT-4o and Qwen \citep{hurst2024gpt}, revealed a significant capability gap. When transitioning from the passive diagnostic task to the active diagnostic task, diagnostic accuracy decreases by 12.75\% on average in our setting. Supporting-evidence quality decreases by 24.36\% under active inquiry. This gap matters because high scores under full-context evaluation may not fully reflect whether a model can elicit and cite discriminative evidence during interactive diagnostic workflows. Specifically, models may output correct diagnoses without consistently eliciting, selecting, or citing the key evidence needed to justify the decision.

\begin{figure*}[h]
\centering
\includegraphics[width=\textwidth]{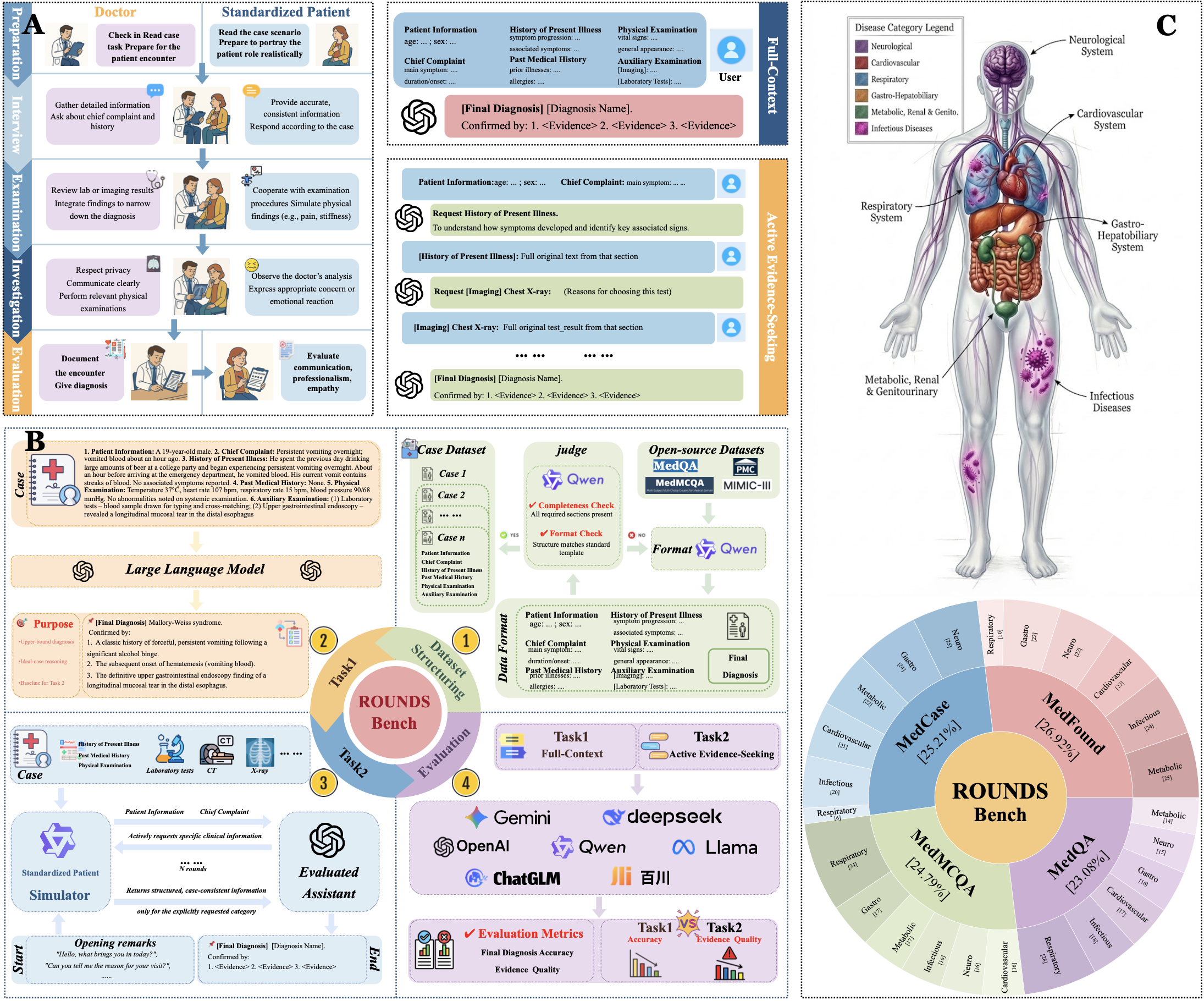} 
\caption{\textbf{Overview of the ROUNDS-Bench framework for evaluating active clinical reasoning.} \textbf{A} Paradigm Comparison. Contrasts the authentic OSCE clinical workflow with the passive Full-Context baseline and the proposed Active Evidence-Seeking interactive protocol. \textbf{B} Technical Pipeline. Illustrates the workflow integrating dataset structuring, dual-task evaluation (Task 1 vs. Task 2), and multi-dimensional grading. \textbf{C} Dataset Stratification. Anatomical visualization of the six clinical systems covered (top) and the distribution of 468 cases across four heterogeneous medical corpora (bottom).
}
\label{fig1}
\end{figure*}

This study not only precisely quantifies this gap but also provides a detailed analysis of typical failure modes in active diagnosis, such as premature diagnostic closure and inefficient information seeking. By identifying these specific weaknesses, we offer critical evaluation benchmarks and optimization directions for building safer and more clinically effective medical AI agents. Through this work, we highlight the limitations of existing medical AI evaluation paradigms and introduce a reproducible interactive evaluation protocol. Our goal is to propel future model evaluation to transition from static question-answering to authentic, dynamic interaction, thereby ensuring usability and safety in clinical decision support applications.

\begin{table}[h]
\centering
\caption{\textbf{Baseline characteristics of the virtual patient cohort in ROUNDS-Bench.} Summary of the 468 standardized clinical cases, stratified by data source (MedQA, MedMCQA, MedFound, and MedCaseReasoning) and six major organ systems to ensure balanced representation and prevent prevalence bias. }
\label{tab1}
\renewcommand{\arraystretch}{1.2} 
\begin{tabular}{lcc}
\toprule
\textbf{Characteristic} & \textbf{No. of Cases} & \textbf{Percentage} \\
 & \textbf{($N=468$)} & \textbf{(\%)} \\
\midrule
\multicolumn{3}{l}{\textbf{Data Source}} \\
\hspace{3mm}MedQA (USMLE-style) & 108 & 23.1 \\
\hspace{3mm}MedMCQA (Indian medical exam) & 116 & 24.8 \\
\hspace{3mm}MedFound (clinical data derived from MIMIC-III) & 126 & 26.9 \\
\hspace{3mm}MedCaseReasoning ( PMC Case Reports) & 118 & 25.2 \\
\midrule
\multicolumn{3}{l}{\textbf{Clinical System Category}} \\
\hspace{3mm}Cardiovascular System & 78 & 16.7 \\
\hspace{3mm}Respiratory System & 78 & 16.7 \\
\hspace{3mm}Gastro-Hepatobiliary System & 78 & 16.7 \\
\hspace{3mm}Neurological System & 78 & 16.7 \\
\hspace{3mm}Infectious Diseases & 78 & 16.7 \\
\hspace{3mm}Metabolic, Renal \& Genitourinary & 78 & 16.7 \\
\bottomrule
\multicolumn{3}{p{0.9\columnwidth}}{\footnotesize \textit{Note:} Percentages may not sum to 100\% due to rounding. The dataset was strictly stratified to ensure equal representation ($N=78$) across all six clinical system categories to prevent prevalence bias during evaluation.}
\end{tabular}
\end{table}

\section{Results}\label{sec2}
In this section, we present our main findings from the evaluation of 15 state-of-the-art LLMs using the ROUNDS-Bench framework. We begin with an overview of the virtual patient cohort and evaluation protocol, followed by a comparative analysis of the capability gap between passive Full-Context diagnosis and Active Evidence-Seeking. Finally, we examine the critical safety risks associated with evidence quality  and dissect the key determinants driving performance variability, including scaling laws, clinical system differences, and data source complexity.

\subsection{The ROUNDS-Bench Framework and Virtual Patient Cohort}
To complement static benchmark settings and quantify interactive evidence acquisition under controlled information release, we developed the ROUNDS-Bench interactive evaluation framework. Distinct from the Full-Context paradigm typified by MedQA or USMLE, ROUNDS-Bench integrates a high-fidelity Standardized Patient Simulator (SP-sim) designed to reproduce the key characteristic of progressive disclosure found in authentic clinical diagnosis. Specifically, the SP-sim incorporates a rigorous Information Gating Mechanism: initiating with only basic demographics and a chief complaint, models must unlock subsequent evidence, such as symptom details, physical exam findings, and lab results, through clinically logical History Taking and Test Requesting. This mechanism mimics the information asymmetry and iterative evidence-seeking process of real-world doctor-patient interactions (detailed in Methods 4.2). By precluding the shortcut of reading the full case summary, this design compels the evaluation to focus squarely on the model's ability to formulate queries, select tests, and dynamically update hypotheses under uncertainty.

To ensure clinical representativeness and cross-domain comparability, we constructed a strictly stratified Virtual Patient Cohort:
\begin{itemize}
    \item \textbf{Clinical Diversity and Balance:} As shown in Table \ref{tab1}, the cohort comprises 468 independent cases, precisely equilibrated across six core clinical systems: Gastro-Hepatobiliary, Neurological, Infectious Diseases, Metabolic, Renal \& Genitourinary, Cardiovascular, and Respiratory, with each system accounting for 16.7\% ($N=78$) of the dataset. This balanced sampling strategy significantly mitigates the impact of disease spectrum bias, ensuring fair comparisons across identical task difficulties and distributions.
    \item \textbf{Data Heterogeneity:} Cases were synthesized and structured from four complementary, high-quality corpora to ensure heterogeneity: MedQA (USMLE-style), MedMCQA (high-difficulty Indian medical exams), MedFound (clinical data derived from MIMIC-III), and MedCaseReasoning (PMC Case Reports)\citep{jin2021disease, pal2022medmcqa, liu2025generalist, wu2025medcasereasoningevaluatinglearningdiagnostic, johnson2016mimic}. This multi-source fusion spans a broad clinical spectrum from typical common presentations to atypical complex conditions, reducing the risk of domain overfitting and enhancing robustness to diverse clinical expressions.
\end{itemize}

Leveraging this framework, we implemented a paired dual-task experimental design to isolate and quantify the capability variance between passive reading and active acquisition workflows:
\begin{itemize}
    \item \textbf{Task 1 (Full-Context):} Models receive the complete electronic health record instantaneously, establishing the theoretical upper bound of static diagnostic proficiency.
    \item \textbf{Task 2 (Active Evidence-Seeking):} Models are provided only with the Chief Complaint and must autonomously drive a multi-turn diagnostic dialogue to progressively acquire evidence. By comparing performance on the same cases across these two settings, we precisely characterize the performance degradation when transitioning from static, full-info diagnosis to active, evidence-based diagnosis, a metric we define as the Capability Gap (Fig. \ref{fig1}).
\end{itemize}

\begin{sidewaystable}[p] 
\centering
\scriptsize 
\setlength{\tabcolsep}{4pt} 
\renewcommand{\arraystretch}{1.3} 

\caption{\textbf{ Leaderboard of clinical reasoning performance across 15 LLMs stratified by parameter scale.} Models are hierarchically ranked within four tiers based on parameter density. This leaderboard contrasts the static performance ceiling (Task 1: Full-Context) with the interactive reasoning capability (Task 2: Active Seeking). The gap metric quantifies the performance erosion inherent in active inquiry, while "StrictEQ" serves as a rigorous audit of evidence grounding. Best-in-class performers within each tier are highlighted in bold, identifying the most clinically efficient architectures under varying computational constraints.}
\label{tab2}

\begin{tabular}{llcccc}
\toprule
\textbf{Model Family} & \textbf{Model Name} & \makecell[c]{\textbf{Task 1: Full-Context} \\ \textit{Exact Accuracy (\%)}} & \makecell[c]{\textbf{Task 2: Active Seeking} \\ \textit{Exact Accuracy (\%)}} & \makecell[c]{\textbf{The Gap} \\ \textit{($T2 - T1$)}} & \makecell[c]{\textbf{Evidence Quality} \\ \textit{StrictEQ (T2, \%)}} \\
\midrule

\multicolumn{6}{l}{\textit{\textbf{Proprietary / Large Models ($>32$B)}}} \\
\hspace{2mm}Gemini & Gemini-2.5-Pro & \textbf{65.2} & \textbf{49.4} & -15.8 & 42.5 \\
\hspace{2mm}DeepSeek & DeepSeek-v3-250324 & 59.2 & 46.2 & -13.0 & \textbf{50.2} \\
\hspace{2mm}OpenAI & ChatGPT-4o-mini & 45.5 & 36.5 & \textbf{-9.0} & 44.0 \\
\midrule

\multicolumn{6}{l}{\textit{\textbf{Medium Models (32B)}}} \\
\hspace{2mm}Qwen & Qwen3-32B & \textbf{59.4} & \textbf{48.3} & -11.1 & \textbf{51.3} \\
\hspace{2mm}Qwen & Qwen2.5-32B-Instruct & 49.4 & 39.1 & \textbf{-10.3} & 49.2 \\
\hspace{2mm}ZhipuAI & GLM-4-32B & 47.2 & 35.7 & -11.5 & 39.5 \\
\hspace{2mm}DeepSeek & DeepSeek-R1-Distill-Qwen-32B & 53.4 & 33.9 & -19.5 & 31.7 \\
\midrule

\multicolumn{6}{l}{\textit{\textbf{Small Models (14B)}}} \\
\hspace{2mm}Qwen & Qwen3-14B & \textbf{57.7} & \textbf{45.1} & -12.6 & \textbf{52.1} \\
\hspace{2mm}Baichuan & Baichuan-M1-14B & 53.4 & 44.9 & \textbf{-8.6} & 44.9 \\
\hspace{2mm}Qwen & Qwen2.5-14B-Instruct & 42.5 & 34.0 & \textbf{-8.6} & 39.5 \\
\hspace{2mm}DeepSeek & DeepSeek-R1-Distill-Qwen-14B & 47.0 & 23.3 & -23.7 & 16.2 \\
\midrule

\multicolumn{6}{l}{\textit{\textbf{Very Small Models (7B-8B)}}} \\
\hspace{2mm}Qwen & Qwen3-8B & \textbf{53.4} & \textbf{39.5} & -13.9 & \textbf{45.3} \\
\hspace{2mm}Meta & Llama-3-8B-Instruct & 36.8 & 24.8 & -12.0 & 22.9 \\
\hspace{2mm}Qwen & Qwen2.5-7B-Instruct & 37.0 & 23.3 & -13.7 & 19.2 \\
\hspace{2mm}DeepSeek & DeepSeek-R1-Distill-Qwen-7B & 14.1 & 6.0 & \textbf{-8.1} & 1.9 \\

\bottomrule
\end{tabular}
\end{sidewaystable}

\begin{figure*}[t]
\centering
\includegraphics[width=\textwidth]{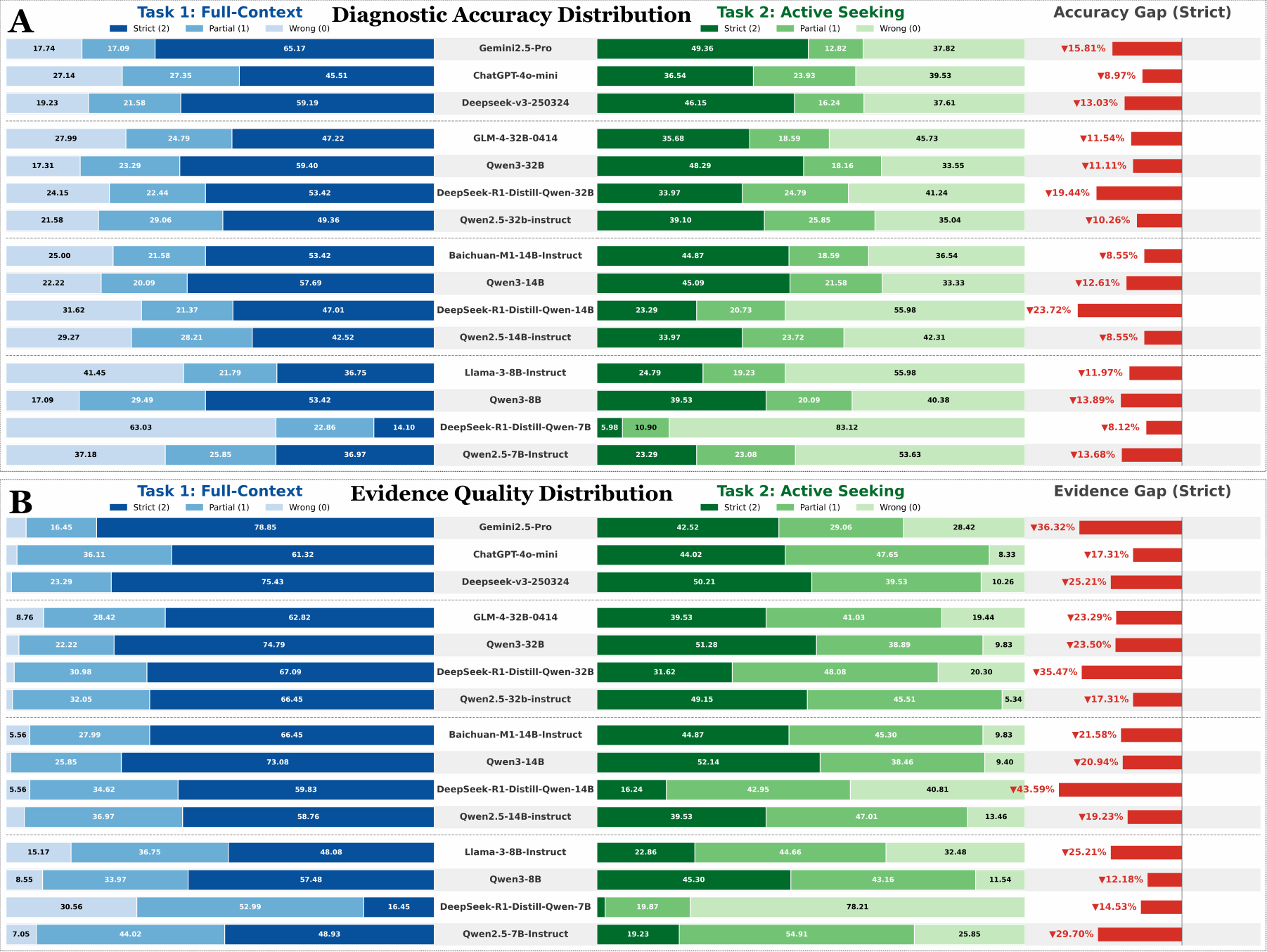} 
\caption{
    \textbf{Distribution of diagnostic outcomes and quantification of the capability gap.} The figure presents a paired analysis of \textbf{A} Diagnostic Accuracy and \textbf{B} Evidence Quality across 15 LLMs. \textbf{Center:} Models are listed centrally, stratified by parameter size (separated by dashed lines). \textbf{Left (Blue panels):} Performance distribution in the static Task 1 (Full-Context) setting. \textbf{Right (Green panels):} Performance distribution in the dynamic Task 2 (Active Evidence-Seeking) setting. Darker shades indicate strict correctness, while lighter shades represent partial correctness or errors. Note the visible expansion of the Wrong (lightest shade) segments in Task 2. \textbf{Far Right (Red bars):} The Performance Drop ($\Delta$) column explicitly visualizes the decline in strict accuracy/evidence quality when transitioning from Task 1 to Task 2. 
  }
\label{fig2}
\end{figure*}

\subsection{Significant Capability Gap in Active Diagnosis}
The transition from the static baseline (Task 1) to the dynamic inquiry setting (Task 2) was associated with a consistent and substantial deterioration in diagnostic performance across all evaluated models. This finding systematically shows a consistent performance drop across models in our interactive setting.\\
\textbf{Quantitative Performance Degradation.} As evidenced in Table \ref{tab2} and Figure \ref{fig2}, the average Exact Accuracy across the full model cohort declined by approximately 12.75\% when transitioning from Task 1 to Task 2. This trend demonstrates a consistent deterioration across diverse architectures when the cognitive load shifts from passively receiving complete case details to actively planning inquiry paths and acquiring evidence.\\
\textbf{Proprietary Models.} Even proprietary models that excelled in the static baseline failed to maintain their superiority in the active diagnostic setting. For instance, Gemini-2.5-Pro, which achieved an Exact Accuracy of 65.20\% in Task 1 among the higher scores observed in the full-context setting in our evaluation, saw its performance retract to 49.36\% in Task 2, an absolute decline of 15.84 percentage points. Similarly, ChatGPT-4o-mini, despite a moderate static baseline (45.51\%), suffered further degradation to 36.54\% (a drop of 8.97 percentage points), exhibiting significant capability erosion.\\
\textbf{Open-Weights Models.} The capability gap was even more pronounced within open-weights architectures. For example, Qwen3-32B declined from 59.40\% in Task 1 to 48.29\% in Task 2 ($\Delta$=11.11\%). Most notably, the reasoning-distilled model DeepSeek-R1-Distill-Qwen-32B experienced the largest performance decline, decreasing from 53.42\% to 33.92\%, a reduction of 19.50 percentage points. These results collectively indicate that significant performance bottlenecks persist for both proprietary and open-weights models when facing tasks requiring autonomous inquiry planning and evidence acquisition.\\
\textbf{Error Distribution Shift.} The chart in Figure \ref{fig2} further elucidates the shifting error patterns underlying this performance decline. Crucially, cases deemed Strictly Correct in Task 1 did not merely degrade into Partially Correct (ambiguous) diagnoses in Task 2; rather, a substantial proportion shifted directly to the Wrong category. This phenomenon, visualized by the significant expansion of the light green region, implies that in dynamic environments with incomplete information, models do not adopt a more conservative or cautious decision-making strategy. Instead, they are prone to constructing erroneous causal chains based on fragmented or insufficient evidence, leading to a complete deviation from the actual disease course. This error distribution shift highlights that the challenge in active diagnosis is not merely a drop in overall accuracy, but a larger share of cases shift directly from strictly correct to wrong in Task 2, rather than to partially correct, suggesting that errors under incomplete information are not consistently mitigated by more conservative outputs.

\subsection{Evidence Quality and Safety Risks: Hallucinated Reasoning}
We observe that the decline in evidence quality (StrictEQ) is larger than the decline in diagnostic accuracy. StrictEQ extends beyond mere diagnostic correctness; it evaluates whether the model can provide a complete, traceable chain of evidence, comprising both key positive and negative findings, derived from actual information acquired during the interaction. As such, StrictEQ serves as a direct proxy for the interpretability and auditability of the decision-making process. Consequently, the observation of a pronounced decoupling between diagnostic accuracy and evidence quality constitutes a distinct safety signal.\\
\textbf{Hallucinated Reasoning.} As illustrated by the prominent red bars in the right panel of Figure \ref{fig2}, the Evidence Gap (the StrictEQ differential between Task 1 and Task 2) is consistently wider than the Accuracy Gap across the majority of models. In other words, as models transition from static to dynamic environments, their ability to retrieve evidence decays significantly faster than their ability to output correct diagnostic labels. This suggests that in a substantial proportion of cases, models arrive at the correct diagnosis via some correct diagnoses are produced without retrieving or citing all discriminative supporting evidence available under our interaction constraints. We operationally define this mode of correct outcome with missing evidentiary basis as Hallucinated Reasoning, effectively getting the right answer for the wrong (or non-existent) reasons.\\
\textbf{Case Analysis.} Gemini-2.5-Pro exemplifies this risk. While it maintained a diagnostic accuracy of 49.36\% in Task 2, its StrictEQ score lagged significantly at 42.52\%. This inversion implies that in approximately 7\% of cases, the model produced "correct diagnoses" that failed to meet strict evidentiary standards. In these instances, the model's final output aligned with the reference standard, yet the dialogue history lacked the complete evidence set required to substantiate that conclusion. For Clinical Decision Support Systems (CDSS), such unsubstantiated inferences are inherently resistant to physician auditing, posing a risk of amplifying clinical errors if blindly trusted.\\
\textbf{Safety Benchmark.} In contrast, DeepSeek-v3-250324 demonstrated superior clinical rigor. Although its diagnostic accuracy (46.15\%) was marginally lower than Gemini's, its StrictEQ reached 54.30\%, the highest among all evaluated models. This indicates a unique characteristic where evidence retrieval performance surpasses diagnostic accuracy. Even in cases where DeepSeek-v3-250324 missed the precise reference diagnosis, it frequently succeeded in systematically acquiring and presenting key clinical evidence. This evidence-leading (as opposed to guess-leading) trait is vital for CDSS safety. It not only mitigates the latent risks of lucky guesses but also provides a solid, auditable foundation for human clinicians, thereby establishing a more robust trust boundary in human-AI collaboration.

\begin{figure}[htbp]
    \centering

    \begin{subfigure}[t]{0.48\textwidth}
        \centering
        \includegraphics[width=\linewidth]{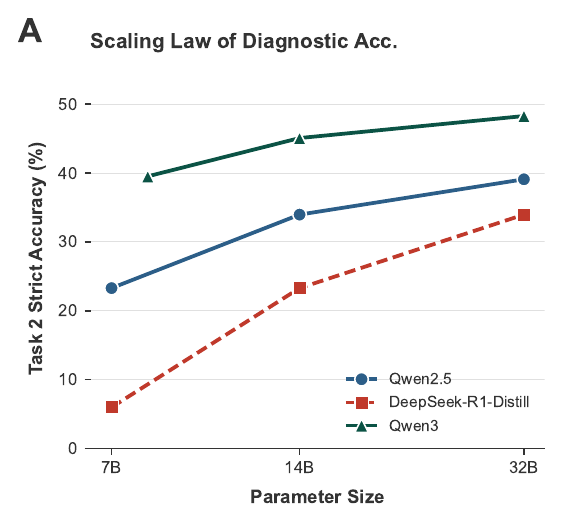}
        \caption{}
        \label{fig:a}
    \end{subfigure}
    \hspace{0.0\textwidth}
    \begin{subfigure}[t]{0.48\textwidth}
        \centering
        \includegraphics[width=\linewidth]{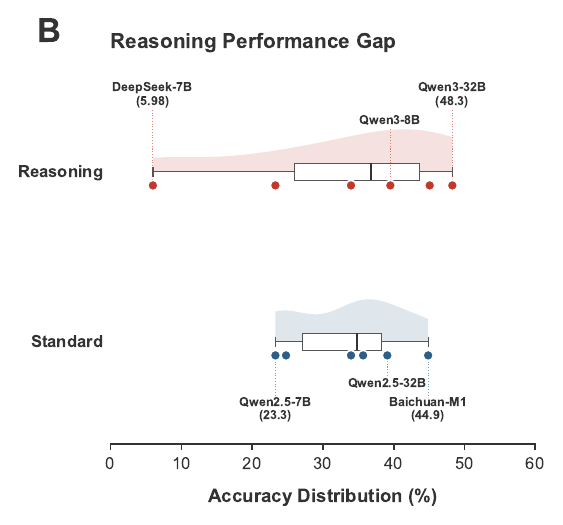}
        \caption{}
        \label{fig:b}
    \end{subfigure}
    \par\vspace{-8pt}
    \begin{subfigure}[t]{0.95\textwidth}
        \centering
        \includegraphics[width=\linewidth]{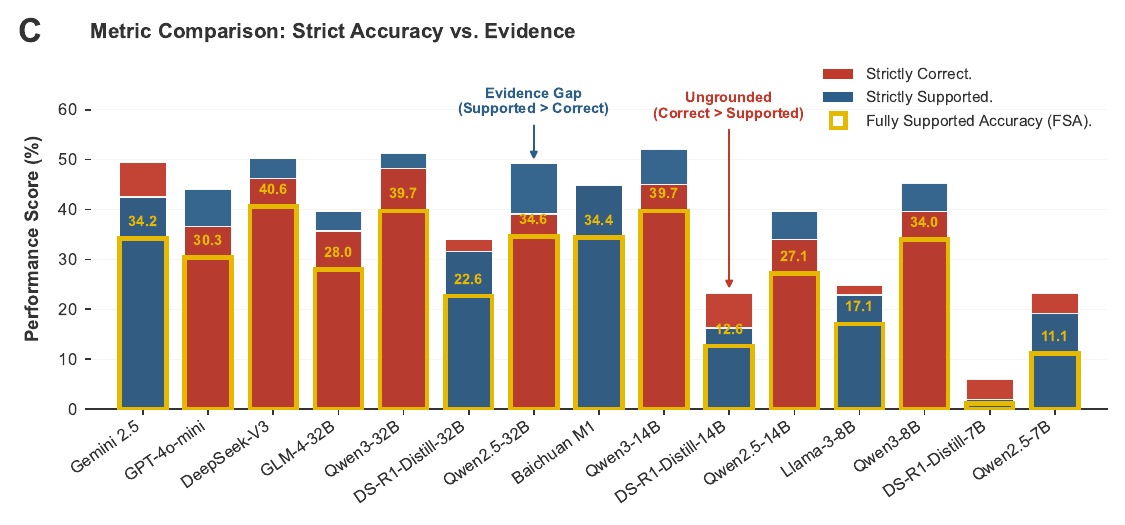}
        \caption{}
        \label{fig:c}
    \end{subfigure}
    \vspace{-8pt}
    \caption{\textbf{Scaling behavior, reasoning architectures, and diagnosis–evidence alignment in Task~2 active diagnosis.} \textbf{A} Strict diagnostic accuracy increases approximately log-linearly with model size within each family, but a residual capability gap remains. \textbf{B} Reasoning-enhanced models are generally more robust than size-matched instruction-tuned baselines, while small distilled reasoning models perform poorly in active inquiry. \textbf{C} For each model, red bars show strict diagnostic accuracy, blue bars Strict Evidence Quality, and gold bars Fully Supported Accuracy (FSA); the shaded “Ungrounded” region marks correct diagnoses lacking fully grounded evidence, whereas the “Evidence-leading” region highlights models whose evidence quality exceeds their diagnostic accuracy.}

    \label{fig3}
\end{figure}

\subsection{Determinants of Performance: Scaling Laws and Reasoning Architecture}
After establishing the existence of a substantial capability gap in active diagnosis, we next examined how model scale, reasoning paradigm, and the alignment between diagnostic accuracy and evidentiary support jointly shape performance (Fig.\ref{fig3}).\\
\textbf{Scaling behavior in active diagnosis.} As shown in Fig.\ref{fig3} A, Task~2 strict diagnostic accuracy exhibits a monotonic, approximately log-linear improvement with increasing parameter size within each model family. Larger models benefit from enhanced multi-turn state tracking and long-context retention, which partially buffer the degradation observed when transitioning from passive to active diagnosis. For example, within the Qwen3 family, strict accuracy in Task~2 rises steadily from the 7B to the 32B variant, with a concomitant reduction in variance across cases. A similar trend is observed for the Qwen2.5 series. These scaling curves indicate that classical scaling laws extend beyond static benchmarks to confer robustness in interactive settings; however, even the largest models still experience a substantial active–passive gap, suggesting that increased capacity alone is insufficient to close the deficit.\\
\textbf{Reasoning architectures and robustness.} Fig.\ref{fig3} B contrasts Task~2 performance between reasoning-enhanced models and size-matched standard instruction-tuned baselines. At medium and large scales, reasoning-oriented architectures such as Qwen3-32B achieve some of the highest strict accuracies in active inquiry, indicating that native long-horizon reasoning can translate into improved diagnostic planning under uncertainty. In contrast, distilled reasoning models at smaller scales, such as DeepSeek-R1-Distill-7B and 14B, display pronounced fragility: their Task~2 accuracies cluster near the bottom of the distribution despite exhibiting competitive performance in static settings. This asymmetry suggests that reasoning capabilities distilled from static Chain-of-Thought traces primarily optimize step-by-step deduction within fully observed contexts and do not consistently generalize to proactive evidence seeking. In other words, these models are proficient at reasoning over given information but struggle to decide what to ask next when information is missing.\\
\textbf{Alignment between diagnosis and evidence.} Beyond raw accuracy, Fig.\ref{fig3} C analyzes how often diagnostic predictions are backed by fully grounded evidence. For each model, the red bar denotes strict diagnostic accuracy in Task~2, the blue bar denotes Strict Evidence Quality (proportion of cases with evidence score $s^{(i)}_{\mathrm{eq}}=2$), and the bright gold overlay corresponds to Fully Supported Accuracy (FSA), defined as the proportion of cases where both the diagnosis and the supporting evidence achieve the maximum score ($s^{(i)}_{\mathrm{acc}}=2$ and $s^{(i)}_{\mathrm{eq}}=2$;). The upper shaded region labeled “Ungrounded” highlights cases in which strict diagnostic accuracy exceeds evidence quality, indicating correct labels that are not fully supported by acquired evidence. Conversely, the “Evidence-leading” region marks cases where evidence quality exceeds strict accuracy, reflecting models that collect coherent evidence but still fail to commit to the correct diagnosis. Clinically, the most desirable pattern is a high FSA with minimal ungrounded mass. In our cohort, only a small subset of models, such as DeepSeek-v3-250324 and Qwen3-32B, begin to approximate this evidence-leading profile, whereas many others display sizable ungrounded segments, reinforcing the safety concerns around hallucinated or unjustified reasoning in active diagnostic workflows.

\begin{figure*}[t]
\centering
\includegraphics[width=\textwidth]{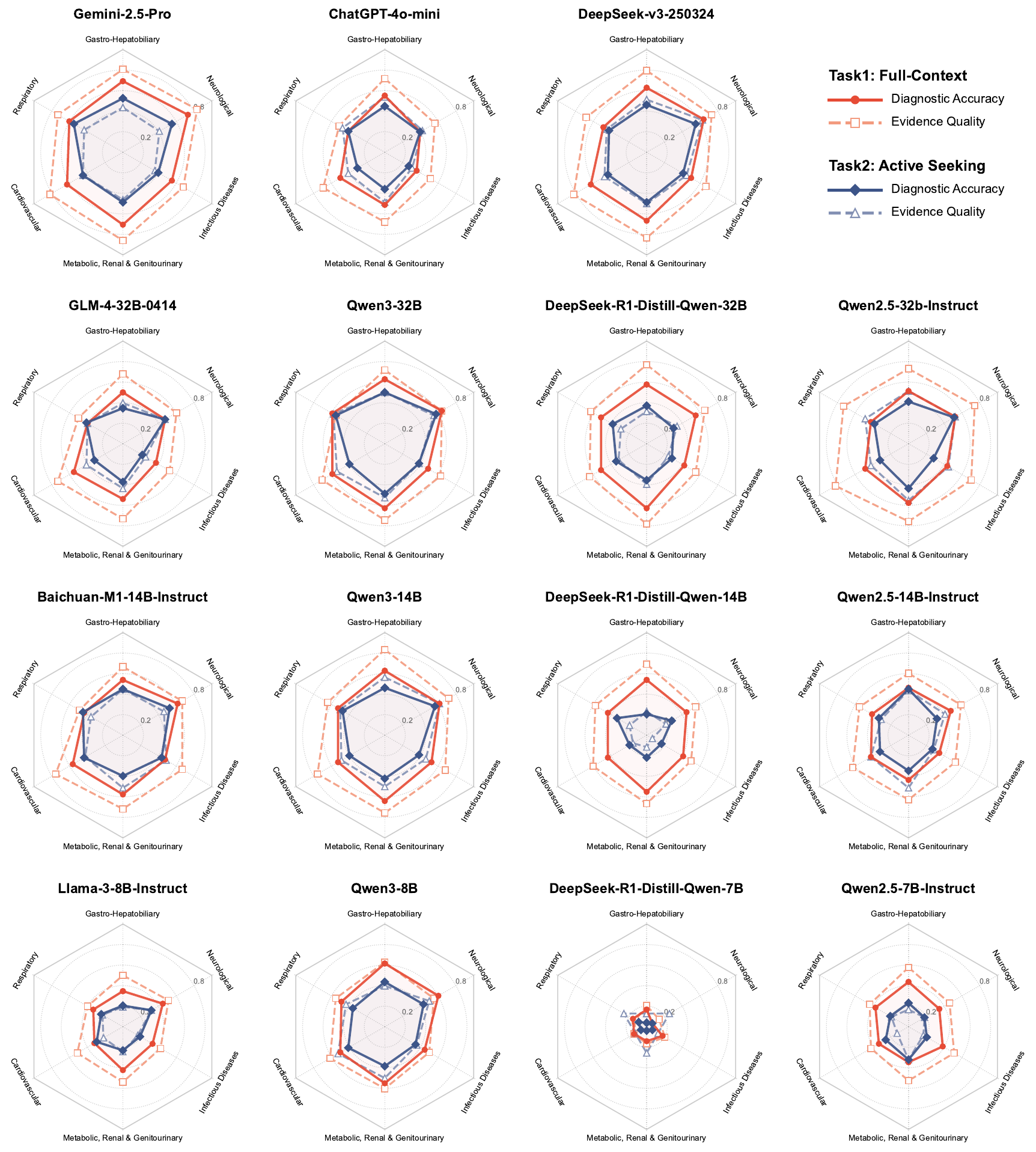} 
\caption{
    \textbf{Specialty-specific performance and diagnostic fragility across clinical systems.} Paired analysis of diagnostic accuracy and evidence quality for each model across six clinical specialties. 
  }
\label{fig4}
\end{figure*}

\subsection{System-Specific Performance Variability}
Performance in active inquiry tasks exhibited pronounced domain-specific variability, a phenomenon intrinsically linked to the distinct diagnostic logic and epistemic structures underlying different clinical specialties (Fig. \ref{fig4}).\\
\textbf{Resilient Domains.} Most models demonstrated robust performance and minimal capability gaps within the Respiratory and Cardiovascular systems. We attribute this resilience to the fact that common pathologies in these fields, such as pneumonia and coronary artery disease, typically follow highly standardized symptom-sign-test-diagnosis mapping pathways. Pathognomonic features, such as specific radiation patterns of chest pain or descriptions of rales possess high textual expressivity and are readily accessible via semi-structured history taking. Consequently, models can effectively capture key diagnostic signals through standard inquiry workflows, keeping performance degradation within a controllable range.\\
\textbf{Vulnerable Domains.} Conversely, models displayed significant fragility in Neurological and Metabolic Renal cases, suffering the steepest declines in both accuracy and evidence quality. This vulnerability stems from specific clinical exigencies: 
    \textbf{Neurological System:} Diagnosis relies heavily on nuanced, structured physical examination, including grading of muscle power and reflexes, assessment of ataxia, cranial nerve deficits, and sensory level localization, and the systematic exclusion of pertinent negative signs. In a text-only interface devoid of multi-modal sensory input, models struggle to execute the precise verification required for anatomical localization. \textbf{Metabolic \& Renal Systems:} These diagnoses necessitate the interpretation of complex combinatorial logic regarding laboratory values, including patterns of electrolyte imbalance, acid–base parameters, and longitudinal trends in renal function. This demands not only numerical sensitivity but also sophisticated longitudinal integration of multi-parametric data. Failure to proactively and precisely request specific biochemical markers often leads to misjudgment of disease severity or etiology.\\
\textbf{Implication.} We observe particularly low performance for some models in neurological differential diagnosis under our text-only protocol. This suggests that domains requiring fine-grained physical exam localization or multi-parameter lab interpretation may be more challenging in this setting, motivating cautious deployment and targeted optimization.

\begin{figure*}[t]
\centering
\includegraphics[width=\textwidth]{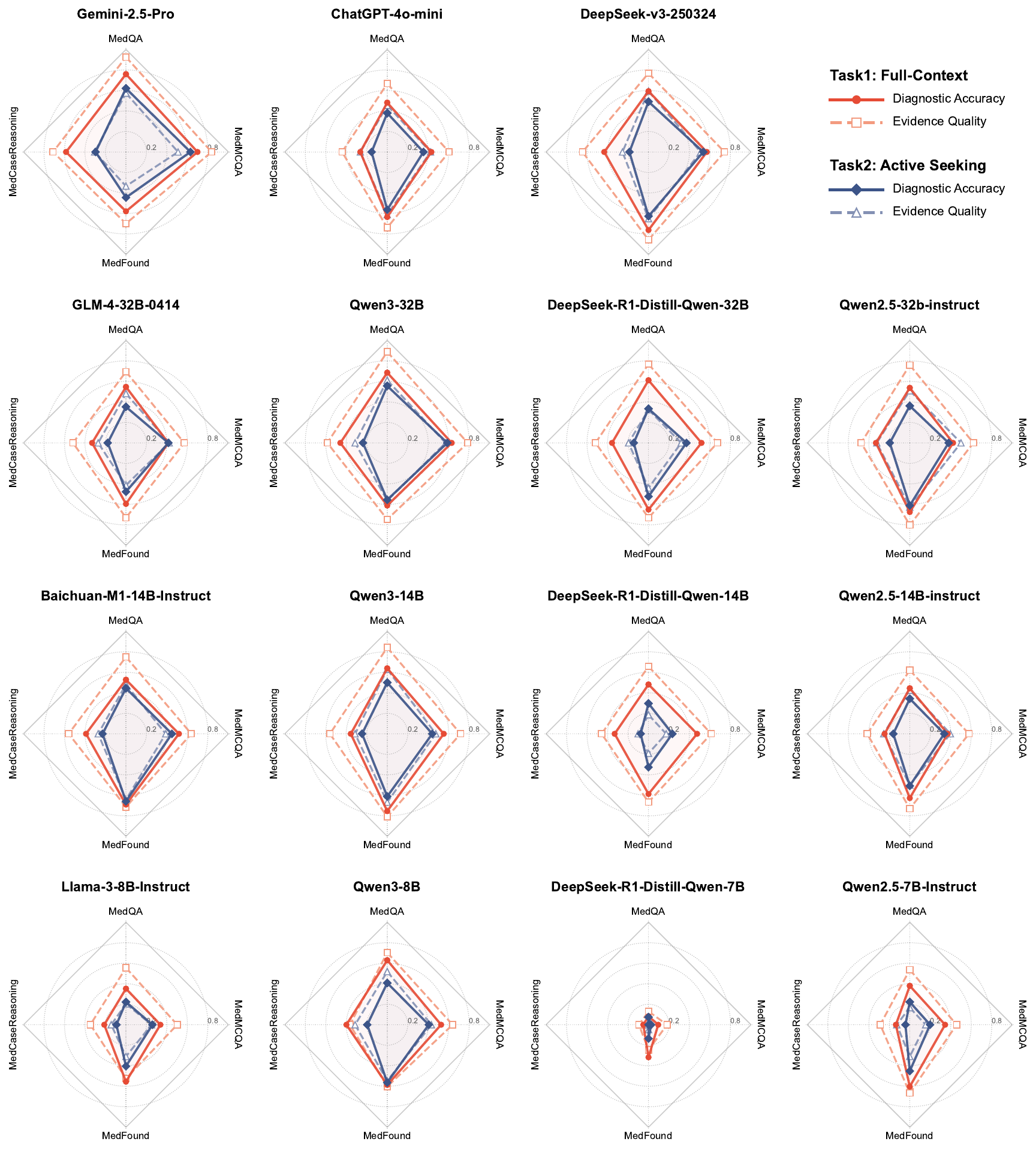} 
\caption{
    \textbf{Impact of data source complexity and information density on active reasoning.}
  }
\label{fig5}
\end{figure*}

\subsection{Impact of Data Source Complexity on Diagnostic Performance}
Beyond inherent variations in clinical systems, the provenance and textual characteristics of case data serve as critical determinants of active diagnostic difficulty. Stratified analysis of Task 2 results reveals a stark performance divergence when models process standardized examination questions versus authentic real-world case reports. (Fig. \ref{fig5}).\\
\textbf{Higher performance on standardized exam-style cases than on case-report-style records.} In cases derived from MedQA (USMLE-style) and MedMCQA, the majority of models maintained elevated diagnostic accuracy in active inquiry tasks (averaging $>50\%$). We attribute this to the highly distilled nature of exam data, which exhibits typical textbook-like characteristics: standardized symptom descriptions, short logical chains, and minimal irrelevant noise. Furthermore, given that models likely encountered similar schemas during pre-training, they can often lock onto diagnoses via pattern matching rather than de novo clinical reasoning. This simplification effect leads to a systematic inflation of evaluation results on standardized benchmarks.\\
\textbf{Precipitous Decline in Real-World Data.} In sharp contrast, performance across all models suffered a marked decline when evaluating real-world case reports from MedCaseReasoning (average accuracy $<35\%$). These datasets mirror the high-entropy nature of actual practice, featuring long-context unstructured narratives with convoluted timelines, such as multiple visits and evolving disease courses, and a plethora of irrelevant confounders, including detailed social history, family history, or incidental findings. In such complex contexts, models must possess the high-order capability to separate signal from noise within redundant text, a challenge far exceeding that of extracting keywords from finite options in standardized tests.\\
\textbf{Quantitative Evidence and Attribution.} Even DeepSeek-R1, possessing the strongest reasoning capabilities, exhibited a performance gap of approximately 20 percentage points between MedQA and MedCase scenarios. This discrepancy elucidates a critical bottleneck: while current LLMs excel at processing structured knowledge, they struggle profoundly with information extraction and noise filtering amidst the unstructured complexity of real clinical records. These results are consistent with the view that performance on exam-style benchmarks may not fully characterize evidence extraction and noise filtering in more heterogeneous case records, and they motivate interactive evaluation as a complementary setting.

\section{Discussion}
\textbf{Performance gap between full-context diagnosis and active evidence seeking.} Using the ROUNDS-Bench framework, we quantify the performance gap between full-context medical question answering and interactive diagnostic inquiry, showing that strong static performance does not reliably carry over to staged, information-sparse settings. Our results highlight a limitation of prevailing full-context evaluations: models that perform well when all evidence is provided upfront often degrade when they must decide what to ask for and how to update hypotheses across turns. he paired decline in diagnostic accuracy and evidence quality suggests caution in treating licensing-style, full-context scores as stand-alone proxies for interactive clinical decision support behavior. More broadly, these findings underscore that static medical QA and interactive diagnostic workups impose different operational demands on models. Interactive diagnosis additionally requires purposeful information acquisition, uncertainty-aware hypothesis management, and evidence synthesis over multiple turns.\\
\textbf{Cognitive interpretation: from pattern completion to hypothesis-driven inquiry.} From a cognitive perspective, the degradation may reflect a mismatch between next-token prediction optimized for pattern completion and the hypothesis-driven inquiry required in clinical workups. In static, full-context settings, models can often map a complete symptom/evidence set to a diagnosis by leveraging distributional regularities learned during training. By contrast, clinical diagnosis typically follows a hypothetico-deductive process in which clinicians generate hypotheses, seek discriminative evidence, and revise beliefs as new information emerges. Our results suggest that many models are less reliable in the strategic layer of this process, often exhibiting inefficient questioning and premature diagnostic closure. One interpretation is that models capture many medical entities and associations, but still struggle with procedural decision-making such as selecting high, yield next questions/tests and using them to update a differential diagnosis over time.\\
\textbf{Safety implications: evidence gaps and reduced auditability in interactive workflows.} A key safety-relevant observation is ``hallucinated reasoning,'' where diagnostic labels can appear correct even when the interaction fails to elicit sufficient supporting evidence. Across many cases, models produced correct labels despite not having acquired the key positive and negative findings needed to justify those conclusions. Although such outcomes can improve scores on static benchmarks, in clinical decision support they reduce auditability and can elevate risk by weakening a clinician’s ability to verify the rationale. A system operating on probabilistic intuition rather than a verifiable chain of evidence is inherently untrustworthy for physician oversight. One possible explanation is that common optimization targets emphasize final-answer correctness more than evidentiary sufficiency and traceable support in multi-turn settings. To improve safety, future work could explicitly penalize unsupported ``correct'' answers and optimize for evidentiary sufficiency and traceability alongside accuracy.\\
\textbf{Determinants: model scale, reasoning scaffolds, and planning for inquiry.} Our analysis of model scale and reasoning-related training signals provides insight into factors associated with improved performance under active inquiry. While larger models appear more resilient in longer interactions, scaling alone does not eliminate the gap between full-context and staged-information settings. By comparison, models with stronger multi-step reasoning behavior (e.g., CoT-style traces or RL-aligned variants) showed better stability in maintaining evidence quality across turns. These patterns suggest that beyond knowledge capacity, planning, deciding what to ask next and when enough information has been gathered is a major bottleneck. Active evidence seeking requires maintaining an evolving state representation and selecting prospective queries, which is closer to agentic decision-making than single-pass next-token prediction over a fixed context.\\
\textbf{Limitations.} Limitations. This study has limitations that are inherent to simulation-based evaluation. First, while the standardized patient simulator improves reproducibility, it abstracts away sociopsychological factors present in real encounters, which may limit direct generalization to practice. Second, the interaction is text-only and does not capture multimodal evidence (e.g., imaging or waveform data) that is central to many clinical domains.Third, a single-reference diagnosis may not fully reflect real-world comorbidity and diagnostic uncertainty. Finally, because parts of the grading rely on LLM-based evaluation, additional validation with expert clinician adjudication would strengthen the conclusions.\\
ROUNDS-Bench provides a complementary evaluation perspective that emphasizes model behavior under staged information release, in addition to what models know under full-context conditions. The observed performance decline suggests that progress on static QA should be paired with methods that improve interactive evidence seeking, auditability, and safety constraints in workflow-like settings. Future research could explore reinforcement learning from clinician feedback to reward evidence-based reasoning, discourage low-yield inquiries, and promote efficient information acquisition. In addition, evaluation could incorporate stopping criteria that test whether models can decide when sufficient information has been gathered before committing to a diagnosis. Improving dynamic inquiry and decision thresholds is likely necessary for medical AI to move from passive assistance toward more reliable participation in clinical workflows.

\section{Methods}\label{sec11}
This section elaborates on the development of ROUNDS-Bench, the interaction mechanics of the Standardized Patient (SP) simulator, the dual-task evaluation protocol, and the multidimensional metrics used for assessment. Additionally, all related prompt templates and data cleaning scripts are included.

\subsection{Data Curation and Stratification Pipeline}

To construct a high-quality benchmark that bridges clinical representativeness with the precise evaluation of active inquiry capabilities, we designed and implemented a rigorous Three-Stage Construction Pipeline. This pipeline encompasses the entire lifecycle of data curation, from the aggregation of heterogeneous multi-source data to multi-stage structural cleaning and clinical system stratification.\\
\textbf{1. Multi-source Aggregation.} To maximize data diversity and minimize domain bias arising from single sources, we aggregated four high-quality medical corpora, covering the full spectrum from standardized examinations to authentic clinical narratives:

\textbf{MedQA (USMLE) \& MedMCQA:} Representing standardized medical licensing examinations, providing expert-reviewed, typical clinical presentations.

\textbf{MedFound \& MedCaseReasoning:} Providing long-context clinical case reports closer to real-world scenarios. These datasets feature unstructured timelines, redundant information, and atypical symptoms, serving to evaluate model robustness in high-noise environments.\\
\textbf{2. Multi-stage Filtering and Structuring.}  Raw data typically exist as unstructured text blocks or contain non-diagnostic questions. To adapt this data for the SP-sim's Gating Mechanism, we developed a processing protocol comprising three strict stages:

\textbf{Stage I: Diagnostic-only Filtering.} To rigorously isolate diagnostic reasoning from treatment or prognostic tasks, we implemented a dual-filtering mechanism: \textbf{Diagnosis-type Filter:} Determines whether the question requires inferring a disease from clinical manifestations, explicitly precluding best next step in management or pathophysiological mechanism queries. \textbf{Diagnosis-term Filter:} Verifies that candidate answers correspond to distinct Disease Entities, rather than symptoms or test names. Cases were retained only if they passed both checks.

\textbf{Stage II: Schema-constrained Structuring.} Leveraging DeepSeek-v3, we restructured the retained free text into structured records comprising six standardized sections. We utilized a prompt with hard constraints (see Appendix \ref{AppA}, Fig. \ref{figA2}) to enforce the following rules:
\textbf{Section Delineation:} (1) Patient Info, (2) Chief Complaint, (3) History of Present Illness, (4) Past Medical History, (5) Physical Exam, and (6) Auxiliary Exam.
\textbf{Prevention of Data Leakage: }Explicit prohibition of including the final diagnosis or any suggestive conclusions within the input sections.
\textbf{Null Value Handling: }Information not present in the source text must be explicitly marked as None, strictly prohibiting the model from generative completion to mitigate hallucination risks.

\textbf{Stage III: Structural Validation and Integrity Check.} We introduced an independent, lightweight Validator to audit the structured data. The validator performs a binary assessment on two dimensions: \textbf{Completeness:} Verifying the presence of all six required sections.
\textbf{Faithfulness:} Verifying that structured content is grounded entirely in the source text, checking for confabulated details. Only cases passing this validation were admitted to the final repository, ensuring data veracity and consistency.\\
\textbf{3. Clinical System Stratification.}
To eliminate the impact of class imbalance on evaluation metrics, we adopted a Rule-based System Categorization strategy rather than random sampling. Using medical ontology logic, cleaned cases were automatically mapped to six core clinical systems: Cardiovascular, Respiratory, Gastro-Hepatobiliary, Neurological, Infectious Diseases, and Metabolic, Renal \& Genitourinary. The final ROUNDS-Bench comprises 468 cases, strictly controlled at 78 cases per system ($N=78$). This balanced stratified design (see Table \ref{tab1}) ensures equitable comparison of model performance across different clinical specialties and supports granular, system-specific analysis.

\subsection{Standardized Patient (SP) Simulator Implementation}
A core methodological innovation of this study is the development of a high-fidelity Standardized Patient Simulator (SP-sim) powered by Large Language Models. Unlike traditional static benchmarks, the SP-sim is specifically engineered to simulate the information asymmetry inherent in authentic clinical encounters: the physician is denied full access to information at the outset and must instead acquire evidence incrementally through sequential inquiry and test ordering. To ensure the controllability and reproducibility of the interaction, we eschewed open-ended, free-form chatbot paradigms in favor of a strictly controlled Request-Response Engine. In this system, the doctor model initiates structured requests, and the SP-sim returns finite, gated information based on predefined rules.\\
\textbf{1. Simulator Architecture and Base Model.} The SP-sim is driven by the Qwen2.5-32B-Instruct model. This selection was motivated by the model's stability in instruction following, enabling it to rigorously execute our gating rules without generating boundary-crossing responses or hallucinations. To maximize determinism and experimental reproducibility, we fixed the inference temperature at 0.0 and locked the random seed, ensuring that identical inputs yield strictly identical outputs from the SP-sim.\\
\textbf{2. Initialization and Opening Strategy.} To prevent doctor models from overfitting to a single opening pattern, we designed 15 diverse opening templates for the SP-sim (see Appendix \ref{AppB}, Fig. \ref{figB1}), ranging from "What brings you in today?" to "Please describe any discomfort." At Turn 0, regardless of the doctor model's opening approach, the SP-sim is programmed to release only two sections from the structured record: (1) Patient Information and (2) Chief Complaint. All other clinical evidence modules, History of Present Illness, Past Medical History, Physical Exam, and Auxiliary Exams, remain in a Hidden State, completely invisible to the doctor model. These sections are unlocked only when the doctor model issues compliant queries in subsequent turns.\\
\textbf{3. Information Gating Mechanism.} The Information Gating Mechanism is the core logic of the SP-sim, designed to strictly control the pace of evidence disclosure while maintaining realism. In each turn, the simulator parses the doctor model's questions in real-time and makes decisions based on natural language understanding and matching rules.\\
\textbf{Allowed Query Scopes.} The SP-sim responds exclusively to explicit queries targeting specific modules: (1) HPI, (2) PMH, (3) Physical Exam, and (4) specific categories of Auxiliary Exams. Generalized queries falling outside these scopes, for example "Tell me everything", do not trigger information release, thereby preventing models from bypassing the sequential inquiry process via unrestricted omniscient queries.\\
\textbf{Intent Recognition and Verbatim Retrieval.} When the doctor model issues a compliant query, such as "Please auscultate the chest", the SP-sim locates the corresponding module in the hidden structured data and executes the following strategy: \textbf{Hit:} If a corresponding entry exists in the structured record, for instance when the Ground Truth records "bibasilar crackles", the SP-sim performs Verbatim Retrieval, returning the raw text without semantic rewriting or summarization. This design ensures that the information exposed in Task 2 is semantically and linguistically identical to the full case record in Task 1, maintaining strict consistency.\textbf{ Miss / Negative Feedback:} If the queried test was not performed or the information is absent from the record, the SP-sim returns a standardized negative response, such as "This test was not performed yet" or "The patient denies this symptom". This fixed-format response serves as a guardrail against hallucination, preventing the SP-sim from fabricating false positives while clearly signaling to the doctor model that the information is absent or negative.\\
\textbf{4. Leakage Prevention and Termination Criteria.} \\
\textbf{Leakage Prevention.} To prevent models from acquiring information beyond the task scope via prompt injection or leading questions, the SP-sim is embedded with high-priority system-level anti-leakage instructions. Regardless of how the doctor model guides the conversation (including direct requests for the diagnosis), the SP-sim is constrained to never reveal the Final Diagnosis or provide suggestive conclusions during intermediate turns, ensuring the independence and rigor of the evaluation.\\
\textbf{Termination Criteria.} The interaction proceeds until one of the following conditions is met:
\textbf{Doctor-Initiated Termination.} The doctor model deems the evidence sufficient and voluntarily outputs the [Final Diagnosis] tag. In our protocol, this marks the decision point for diagnostic judgment.
\textbf{Maximum Turn Limit Reached.} If the dialogue reaches the preset maximum (Max Turns = 10), the SP-sim ceases to provide further information, and the doctor model is forced to generate a final diagnosis based on currently available evidence. This constraint prevents infinite inquiry loops and incorporates the ability to judge "when to stop" into the evaluation dimensions.

\subsection{Evaluation Framework}
To precisely quantify the impact of active evidence-seeking on diagnostic performance, we designed a rigorous within-subject controlled dual-task experiment. The core objective of this protocol is to methodologically isolate and contrast model capabilities in static knowledge retrieval versus dynamic information acquisition and reasoning, while holding the underlying clinical case content constant.

\textbf{Task 1: Full-Context Diagnosis (Static Baseline.)} Serving as the control condition, Task 1 evaluates the model's diagnostic upper bound under conditions of Information Transparency.
\begin{itemize}
    \item \textbf{Input:} The model receives the fully structured and cleaned case record $\mathcal{X}_{\text{full}}$ instantaneously. This includes all available history, physical examination details, and auxiliary test results, exposing all diagnostic evidence in a single pass.
    \item \textbf{Output Requirements:} The model is instructed to directly generate a [Final Diagnosis] and concurrently list three key evidence items supporting that diagnosis. Evidence can be drawn from any section without sequential constraints.
    \item \textbf{Objective:} To establish the Theoretical Performance Upper Bound achievable under idealized conditions with zero information acquisition barriers, serving as a reference baseline for measuring performance loss in the active setting.
\end{itemize}

\textbf{Task 2: Active Evidence-Seeking (Dynamic Evaluation).}
Task 2 simulates the entropy reduction process inherent in authentic clinical practice, transitioning from high uncertainty to diagnostic convergence. The model assumes the role of a physician, tasked with actively constructing the patient profile from an initial state of information scarcity through multi-turn interaction.

\begin{itemize}
    \item \textbf{Initialization:} The model is provided only with Patient Demographics and the Chief Complaint, such as "Male, 45, chest pain for 3 hours". All information regarding HPI, PMH, Physical Exam, and Auxiliary Exams remains in a Hidden State, completely invisible to the model.
    \item \textbf{Discrete Action Space ($\mathcal{A}$):} In each turn, the model must execute a single action from a predefined set:
\begin{itemize}
  \item \textbf{History Taking}: Request [History of Present Illness] or [Past Medical History].
  \item \textbf{Physical Exam}: Request [Physical Examination] to retrieve vital signs and physical findings.
  \item \textbf{Diagnostic Testing}: Issue specific orders via Request [Test Name], such as Request [Chest X-ray].
  \item \textbf{Termination}: Output [Final Diagnosis] to conclude the interaction.
\end{itemize}
    \item \textbf{Interaction Constraints and Process Control:} To simulate Resource Constraints and Time Pressure in clinical practice, we impose strict limitations:Single-Step Execution: Only one query is permitted per turn. This constraint compels the model to perform Clinical Triage, prioritizing queries based on diagnostic yield rather than batch-requesting tests to evade decision planning.Turn Cap: The interaction is strictly limited to 10 turns. If this limit is reached, the model is forced to render a diagnosis based on currently available evidence.No Repetition: Redundant requests for previously acquired information or tests marked Not Performed are prohibited to penalize inefficient operational loops.
    \item \textbf{Output Specification \& Evidence Grounding:} Non-Terminal Turns: Models must output in the format (Action): (Rationale), explicitly stating the clinical reasoning behind the chosen action to enhance the interpretability of the inquiry process. Evidence Grounding Rule: In the final diagnostic phase, the three supporting evidence items $\hat{E}$ must be derived strictly from content actually released by the SP-sim during the Interaction History. Citing information that never appeared in the dialogue, such as hallucinated results or data visible only in Task 1, constitutes Hallucinated Evidence and results in a penalty to the Evidence Quality score.
\end{itemize}

\subsection{Evaluation Metrics}
To achieve objective, quantifiable, and reproducible assessment within the context of unstructured natural language generation, we established a Deterministic Evaluation Protocol based on the LLM-as-a-Judge paradigm. This protocol centers on two core endpoints: (1) Diagnostic Accuracy and (2) Reasoning Evidence Quality.\\
\textbf{1. Evaluator Setup \& Protocol.} To ensure consistency across different models and runs, we employed DeepSeek-v3 as a fully independent Arbitrator, implementing strict parameter controls on its inference behavior:
\textbf{Deterministic Inference:} All evaluation calls were executed with a temperature of 0.0 and Top-p of 1.0. This eliminates sampling stochasticity, ensuring that identical inputs yield strictly consistent scores across multiple evaluations. \textbf{Blinded Evaluation:} The arbitrator operates in a blinded manner, accessing only the model's output (diagnosis and evidence) and the Gold Standard. Although DeepSeek-v3 is also included as one of the evaluated models, the evaluator is used in a strictly separate capacity with frozen weights, standardized prompts, and blinded access to model identities, thereby minimizing potential bias toward specific architectures or stylistic patterns.\textbf{Standardized Prompts:} We designed structured scoring prompts to guide the arbitrator based on predefined scales, rather than allowing free-form commentary, to maximize the standardization of the scoring process.\\
\textbf{2. Exact Diagnostic Accuracy.} We prioritize clinical precision over mere semantic proximity. For each case $i$, the model-generated final diagnosis $\hat{y}^{(i)}$ is semantically compared against the gold standard $y^{*(i)}$ using a 3-Level Scoring Scale to derive the score $s_{acc}^{(i)}$:
\begin{itemize}
    \item \textbf{2 (Strictly Correct):} The diagnosis implies Semantic Equivalence with the gold standard, for example Myocardial Infarction vs. Heart Attack, and exhibits precise matching of the Disease Subtype.
    \item \textbf{1 (Partially Correct):} The diagnosis falls within the correct Disease Category or family but lacks specificity, for example Meningitis vs. Viral Meningitis, or presents an incorrect subtype within the correct lineage.
    \item \textbf{0 (Incorrect):} The diagnosis is factually wrong, unrelated, or consists only of non-diagnostic entities, such as symptoms.
\end{itemize}
\textbf{Primary Metric:} We report Exact Diagnostic Accuracy (ExactAcc), defined as the proportion of cases in the total sample $N$ achieving a score of 2:
\begin{equation}
\text{ExactAcc} = \frac{1}{N} \sum_{i=1}^{N} \mathbb{I}[s_{acc}^{(i)} = 2].
\end{equation}
This metric intentionally excludes "partially correct" instances, establishing a rigorous standard for diagnostic precision aligned with clinical requirements.\\
\textbf{3. Evidence Quality (StrictEQ).} To quantify the risk of "Hallucinated Reasoning", where models "guess" the correct diagnosis without reliable evidentiary support, we introduced the Strict Evidence Quality (StrictEQ) metric. This metric specifically evaluates whether the model can construct a complete, traceable chain of evidence based only on information actually acquired during the interaction. For each case, the model outputs three evidence items $\hat{E} = \{e_1, e_2, e_3\}$. The assessment involves two constraints:\\
\textbf{Grounding Check:} In Task 2, we first extract the set of all information fragments actually retrieved by the model during $T$ turns from the SP-sim interaction log $\mathcal{S}_T$. We then verify whether each claimed evidence item $e_k$ can be semantically matched within this history:
\begin{equation}
\forall \hat{e}_k \in \hat{E},\ \hat{e}_k \in \mathrm{SemanticMatch}(S_T).
\end{equation}
If an evidence item $e_k$ cannot be sourced from the interaction log, meaning the test was never requested, the sign was never disclosed, or the model fabricated a "new fact", it is classified as "Hallucinated Evidence." This step strictly distinguishes between citations grounded in real interaction and confabulated details.\\
\textbf{Scoring Criteria:} Upon passing the grounding check, evidence is further
scored based on logical support strength \(s_{\text{eq}}^{(i)} \in \{0,1,2\}\):
At the dataset level, we summarize strict evidence performance using the
Strict Evidence Quality (StrictEQ) metric, defined as the proportion of cases
with maximally supported evidence:
\begin{equation}
\text{StrictEQ} = \frac{1}{N} \sum_{i=1}^{N} \mathbb{I}\big[s_{\text{eq}}^{(i)} = 2\big].
\end{equation}
\begin{itemize}
    \item \textbf{2 (Strictly Supported):} All three evidence items are grounded in the interaction history; and the evidence provides clear, positive clinical support for the diagnosis, such as pathognomonic signs or exclusionary negatives.
    \item \textbf{1 (Partially Supported): }Only 1–2 items are valid, or the logical link to the diagnosis is weak/tangential.
    \item \textbf{0 (Unsupported):} All evidence is hallucinated, contradicts the history, or lacks a causal relationship with the diagnosis.
\end{itemize}
\textbf{Fully Supported Accuracy (FSA).} While ExactAcc captures how often the final diagnosis is strictly correct, it says
nothing about whether the accompanying evidence is reliable. To jointly require
both perfect diagnosis and perfect evidentiary support within the same task, we
define the Fully Supported Accuracy (FSA) metric.
At the case level, we refer to \(s_{\text{acc}}^{(i)} = 2\) as a strictly correct
diagnosis and \(s_{\text{eq}}^{(i)} = 2\) as strictly supported evidence. FSA is
then defined as the proportion of cases in the total sample \(N\) that satisfy both
conditions simultaneously:
\begin{equation}
\text{FSA} = \frac{1}{N} \sum_{i=1}^{N}
\mathbb{I}\big[s_{\text{acc}}^{(i)} = 2 \ \wedge\ s_{\text{eq}}^{(i)} = 2\big].
\end{equation}
By construction, FSA is a joint metric that extends beyond accuracy-only measures:
it captures the subset of model predictions that are not only strictly correct but
also fully supported by grounded evidence. Applied separately to Task 1 and Task 2,
FSA enables us to quantify how many clinically safe, well-justified decisions are
preserved when moving from static full-context evaluation to active evidence-seeking.

\subsection{Evaluated Models and Implementation Details}
To comprehensively benchmark the performance of the current LLM ecosystem in active clinical reasoning, we selected a representative cohort of 15 models, stratified by Parameter Scale, Access Mode, and Reasoning Architecture. Our evaluation spans the full spectrum from lightweight edge models to hyperscale cloud-based models:
\begin{itemize}
    \item \textbf{Proprietary SOTA Models:} This tier establishes the performance benchmark and includes Gemini-2.5-Pro, ChatGPT-4o-mini, and DeepSeek-v3-250324. These models represent the current upper bound of generalist intelligence and large-scale architectural optimization\citep{deepseek2024deepseek, team2023gemini, comanici2025gemini}.
    \item \textbf{Reasoning-Enhanced Models:} A core focus of our study is the evaluation of models featuring native Chain-of-Thought (CoT) and long-horizon reasoning capabilities. This category includes the Qwen3 series (8B, 14B, 32B) and DeepSeek-R1-Distill (7B, 14B, 32B)  variants. Unlike standard models, these reasoning models are optimized via Reinforcement Learning (RL) to internalize exploratory reasoning paths, allowing us to assess whether native CoT improves diagnostic inquiry \citep{guo2025deepseek, yang2025qwen3}.
    \item \textbf{Generalist Instruction-Tuned Models:} To provide a controlled comparison, we included the Qwen2.5 series (7B, 14B, 32B), Llama-3-8B-Instruct, GLM-4-32B, and Baichuan-M1-14B. This enables an empirical analysis of the  Reasoning Premium, the performance gain attributable to reasoning specific tuning over standard instruction-following \citep{glm2024chatglm, wang2025baichuan, grattafiori2024llama, hui2024qwen2}.
\end{itemize}

\bmhead{Acknowledgements}
This work was supported by the National Natural Science Foundation of China (82303695), Shanghai Municipal Natural Science Foundation (23ZR1425400), and Shanghai Soft Science Project (25692114700), 2025 Annual Special Program for the ``Public Benefit'' Science Popularization and Innovation Project of Xuhui District,  xhkp-HM-2025037,  the Clinical Research Project of the
Shanghai Municipal Health Commission (20214Y0468), Zhongshan Hospital Management Science Foundation Project 2024ZSGL09 and Zhongshan Hospital Medical Humanities and Ideological Education Research Project 2024YXRWSZ-007. 

\section*{Code availability}
The code used to generate the results in this study is publicly available at \url{https://github.com/Leonard-zc/ROUNDS-Bench}.

\section*{Author contributions}
Chen Zhan, Xihe Qiu, Xiaoyu Tan and Xibing Zhuang contributed equally to this work. Chen Zhan, Xihe Qiu, Xiaoyu Tan and Xibing Zhuang conceived and designed the study. Chen Zhan and Gengchen Ma conducted the experiments and collected the data. Chen Zhan and Yue Zhang analyzed the data. Chen Zhan, Xihe Qiu and Xiaoyu Tan drafted the manuscript. Shuo Li and Peifeng Liu provided critical feedback. Xihe Qiu, Xiaoxiao Ge, Liang Liu and Lu Gan supervised and administered the project, acquired funding, and revised the manuscript. All authors reviewed and approved the final manuscript.

\section*{Competing interests}
The authors declare no competing interests.
\bibliography{sn-bibliography}

\begin{appendices}

\section{Extended dataset construction}\label{AppA}
This appendix provides the exact prompt templates and examples used in the three-stage data curation pipeline described in Section~4.1. We include only implementation details that are necessary for replication and omit conceptual descriptions that already appear in the main text.
\subsection{Stage I: diagnosis-type and diagnosis-term filtering}
To isolate items that genuinely require diagnostic reasoning, we first apply a pair of lightweight filters that remove questions about mechanisms, treatment, prognosis, or next best step decisions. \ref{figA1}

\begin{figure*}[h]
\centering
\begin{minipage}[t]{0.48\textwidth}
\begin{tcolorbox}[
  colback=gray!3,
  colframe=gray!60!black,
  title=\small\textbf{Prompt 1 — Diagnosis-Type Filter},
  fonttitle=\bfseries,
  coltitle=black,
  boxrule=0.4pt,
  arc=1mm,
  left=1mm,
  right=1mm,
  top=0.5mm,
  bottom=0.5mm,
  sharp corners=south,
]
\footnotesize
You are a medical board exam assistant. Please determine whether the following question is a diagnosis-type question.

A diagnosis-type question requires the examinee to determine the most likely diagnosis or disease based on the clinical presentation. It does not ask about etiology, treatment, mechanisms, or next steps.

Please answer with only one word: “Yes” or “No”.

\textbf{Question stem:} \texttt{<question\_text>} \\
\textbf{Options:} \texttt{<options\_text>} \\
\textbf{Answer:}
\end{tcolorbox}
\end{minipage}
\hfill
\begin{minipage}[t]{0.48\textwidth}
\begin{tcolorbox}[
  colback=gray!3,
  colframe=gray!60!black,
  title=\small\textbf{Prompt 2 — Diagnosis-Term Filter},
  fonttitle=\bfseries,
  coltitle=black,
  boxrule=0.4pt,
  arc=1mm,
  left=1mm,
  right=1mm,
  top=0.5mm,
  bottom=0.5mm,
  sharp corners=south,
]
\footnotesize
You are a medical expert. Please determine whether the following terms are medical diagnoses (names of diseases or conditions), rather than symptoms, tests, mechanisms, or treatments.

Please answer with only one word: “Yes” or “No”.

\textbf{Options:} \texttt{<options\_text>}
\end{tcolorbox}
\end{minipage}
\caption{Side-by-side prompts used for diagnostic-only filtering (Stage 1).}
\label{figA1}
\end{figure*}

\subsection{Stage II: schema-constrained structuring}
For all retained items, we convert the free-text vignette into a six-section clinical record using a schema-constrained prompt:\ref{figA2}

\begin{figure*}[t]
\centering
\begin{minipage}{\textwidth}
\begin{tcolorbox}[
  colback=gray!3,
  colframe=gray!60!black,
  title=\small\textbf{Prompt: Schema-Constrained Structuring},
  fonttitle=\bfseries,
  coltitle=black,
  boxrule=0.6pt,
  arc=1mm,
  left=1.5mm,
  right=1.5mm,
  top=1.5mm,
  bottom=1.5mm,
  sharp corners=south,
]
\footnotesize
You are a clinical documentation expert. Convert any form of clinical vignette or patient record into a standardized medical record format. Do NOT infer or include any diagnosis.

Use this exact structure:

\textbf{1. Patient Information} \\
- [Sex, Age] (or ``None'') \\

\textbf{2. Chief Complaint} \\
- [Primary symptom] + [Duration] (or ``None'') \\

\textbf{3. History of Present Illness} \\
- Progression: [Chronological illness course] \\
- Accompanying symptoms: [Comma-separated symptoms] \\
(Use ``None'' if not available) \\

\textbf{4. Past Medical History} \\
- [Relevant history] (or ``None'') \\

\textbf{5. Physical Examination} \\
- Vital signs: [Temperature, HR, RR, BP...] (or ``None'') \\
- [System-specific findings] (or ``None'') \\

\textbf{6. Auxiliary Examination} \\
- (1) Imaging test: [Findings] (or ``None'') \\
- (2) Laboratory tests: [Key abnormal results] (or ``None'') \\
- (3)..... \\

Repeat back only the structured output. \\
Please convert the following clinical vignette or case into a structured medical record. \\
Do NOT include any diagnosis results. Fill "None" for missing fields. The text may be a clinical note or exam question. \\

\textbf{Case:} \texttt{{question\_text}} \\

Please strictly follow this format, but leave out the final diagnosis: \\
\texttt{1.Patient Information} \\
\texttt{2.Chief Complaint} \\
\texttt{3.History of Present Illness} \\
\texttt{4.Past Medical History} \\
\texttt{5.Physical Examination} \\
\texttt{6.Auxiliary Examination}
\end{tcolorbox}
\end{minipage}
\caption{Schema-constrained prompt used to convert free-text vignettes into a six-section clinical record (no diagnosis). The schema enforces fixed headers, no leakage, and explicit "None" for missing fields.}
\label{figA2}
\end{figure*}

\subsection{Stage III: structural validation and fabrication check}
A separate validation prompt audits the structured record for completeness and faithfulness:\ref{figA3}

\begin{figure*}[t]
\centering
\begin{minipage}{\textwidth}
\begin{tcolorbox}[
  colback=gray!3,
  colframe=gray!60!black,
  title=\small\textbf{Prompt: Structure \& Fabrication Check},
  fonttitle=\bfseries,
  coltitle=black,
  boxrule=0.6pt,
  arc=1mm,
  left=1.5mm,
  right=1.5mm,
  top=1.5mm,
  bottom=1.5mm,
  sharp corners=south,
]
\footnotesize
You are a critical evaluator of medical documentation quality.

Your task is to determine whether the following structured medical record is a faithful and properly formatted transformation of the original clinical case description.

\textbf{Evaluation Criteria:}
\begin{enumerate}
  \item The record must contain all 6 required sections with their respective content:
  \begin{itemize}
    \item Patient Information  
    \item Chief Complaint  
    \item History of Present Illness  
    \item Past Medical History  
    \item Physical Examination  
    \item Auxiliary Examination
  \end{itemize}
  \textit{Note: Minor variations in section titles (e.g., spacing, punctuation, casing) are acceptable as long as the structure is clearly preserved.}

  \item The structured record must \textbf{not fabricate any content}. All included details must be:
  \begin{itemize}
    \item Explicitly stated in the original case, \textbf{or}  
    \item Clearly implied with no assumptions beyond clinical description.
  \end{itemize}

  \item If any section lacks source information, using \texttt{"None"} is acceptable.
\end{enumerate}

\textbf{Original Case:} \texttt{{original\_text}} \\
\textbf{Structured Medical Record:} \texttt{{formatted\_record}}

Please assess strictly but reasonably. \\
\textbf{Answer only with} \texttt{yes} (fully valid) \textbf{or} \texttt{no} (any fabrication, omission, or structural failure).
\end{tcolorbox}
\caption{Validation prompt for auditing the structured record against the original text: all six sections present, no fabrication (``None'' allowed), and faithful formatting. Output is a strict yes/no.}
\label{figA3}
\end{minipage}
\end{figure*}

\subsection{Illustrative structured case (no diagnosis).}
Figure~\ref{figA4} presents an example record after structuring: demographics, a concise chief complaint with timeline, focused HPI, vitals and exam, and multi-modal auxiliary studies (imaging, laboratory, ECG, hemodynamics, biopsy).
No diagnostic label is included at this stage, preserving the intended separation between \emph{evidence} and \emph{diagnosis} for both Task~1 (full-context) and Task~2 (active evidence-seeking).

\begin{figure*}[t]
\centering
\begin{minipage}{\textwidth}
\begin{tcolorbox}[
  colback=gray!3,
  colframe=gray!60!black,
  title=\textbf{Structured Case Example},
  fonttitle=\bfseries,
  coltitle=black,
  boxrule=0.6pt,
  arc=1mm,
  left=1.5mm,
  right=1.5mm,
  top=1.5mm,
  bottom=1.5mm,
  sharp corners=south,
]
\footnotesize
\textbf{1. Patient Information}  \\ 
- Male, 44 \\

\textbf{2. Chief Complaint}  \\ 
- Chills for 3 days and arthralgias in the knees and hips (preceded by several days of unproductive cough and headache) \\

\textbf{3. History of Present Illness}  \\ 
- Progression: Unproductive cough and headache preceded chills and arthralgias. One week before presentation, he was treated with a macrolide antibiotic and an NSAID. \\ 
- Accompanying symptoms: Cough, Headache, Chills, Arthralgias \\

\textbf{4. Past Medical History}  \\ 
- Smoking history (None otherwise) \\

\textbf{5. Physical Examination}  \\ 
- Vital signs: Temperature 38.5\,°C, Heart Rate 113/min, Blood Pressure 126/64 mmHg, Oxygen Saturation 98\% on room air \\ 
- Findings: No pericardial rub or crackles; epigastric tenderness \\

\textbf{6. Auxiliary Examination}  \\ 
- (1) Imaging test: Chest radiograph showed mild peribronchial cuffing. Transthoracic echocardiography revealed preserved LV function, a 9-mm pericardial effusion, and slight IVC dilation. Coronary CT excluded obstructive disease. Cardiac MRI demonstrated myocardial edema with multifocal subepicardial and subendocardial late gadolinium enhancement and pericardial inflammation. \\ 
- (2) Laboratory tests: WBC 13.4\,$\times$\,10\textsuperscript{3}/$\mu$L, CRP 16.9 mg/dL, ESR 95 mm/h, high-sensitivity troponin T 656.2 ng/L; differential count showed no eosinophilia. Blood cultures, serology, and PCR for pathogens negative; vasculitis-associated autoantibodies absent. \\ 
- (3) Electrocardiogram: Sinus tachycardia with first-degree AV block (PQ 210 ms). \\ 
- (4) Right heart catheterization: Cardiac Index 1.65 L/min/m\textsuperscript{2}, mean PCWP 34 mmHg, LVEDP 29 mmHg. \\ 
- (5) Endomyocardial biopsy: Eight specimens obtained from the left ventricle.
\end{tcolorbox}
\end{minipage}
\caption{Illustrative example of a structured case produced by the schema in Fig.~\ref{figA2}. The record contains demographics, chief complaint with timeline, focused HPI, exam with vitals, and auxiliary studies. No diagnosis is included.}
\label{figA4}

\end{figure*}

\subsection{System-level categorization.}
Finally, diseases are mapped to six clinical systems using a transparent rule-based prompt (Fig.~\ref{figA5}).
The instruction returns a JSON tuple \{\texttt{primary\_diagnosis}, \texttt{category}\}, aligning each case to one of the six systems or \texttt{Other} when appropriate.
This categorization underpins the balanced distribution reported in Table~\ref{tab1} and supports per-system analyses in the experiments.

\begin{figure*}[t]
\centering
\begin{minipage}{\textwidth}
\begin{tcolorbox}[
  colback=gray!3,
  colframe=gray!60!black,
  title=\textbf{Prompt: System-Level Categorization},
  fonttitle=\bfseries,
  coltitle=black,
  boxrule=0.6pt,
  arc=1mm,
  left=1.5mm,
  right=1.5mm,
  top=1.5mm,
  bottom=1.5mm,
  sharp corners=south,
]
\footnotesize
You are a medical assistant with extensive clinical knowledge. Your task is to classify each disease name into one of the following six major clinical system categories. If a disease does not clearly belong to any of these categories, label it as \textbf{``Other''}. \\

\textbf{Use the following classification rules:} \\

\textbf{1. Cardiovascular System} \\
- Includes: Acute coronary syndrome, heart failure, arrhythmias (e.g., atrial fibrillation), hypertensive emergencies, aortic dissection, pericarditis, etc. \\

\textbf{2. Respiratory System} \\
- Includes: Asthma, COPD, pneumonia, pulmonary embolism, spontaneous pneumothorax, hemoptysis-related diseases, etc. \\

\textbf{3. Gastro-Hepatobiliary System} \\
- Includes: Upper or lower GI bleeding, appendicitis, cholecystitis, pancreatitis, liver cirrhosis and complications, inflammatory bowel disease, abdominal pain, diarrhea, etc. \\

\textbf{4. Neurological System} \\
- Includes: Ischemic stroke, TIA, seizures/epilepsy, subarachnoid hemorrhage, headaches, dizziness/vertigo, migraine, CNS infections, etc. \\

\textbf{5. Infectious Diseases} \\
- Includes: Bacterial meningitis, urinary tract infections, community or hospital-acquired pneumonia, skin and soft tissue infections, early sepsis, tropical diseases (e.g., dengue, malaria), etc. \\

\textbf{6. Metabolic, Renal \& Genitourinary System} \\
- Includes: Diabetes mellitus (DKA, HHS), hypoglycemia, thyroid disorders (hyper/hypothyroidism), electrolyte disorders (e.g., hyponatremia, hyperkalemia), acute kidney injury, kidney stones, urinary tract diseases, etc. \\

If a disease does not fit into any of the above six categories, classify it as: \textbf{Other} \\

Please return your answer in JSON format, like this: \\
\texttt{\{"primary\_diagnosis": "Pneumonia", "category": "Respiratory System"\}}
\end{tcolorbox}
\end{minipage}
\caption{Rule-based categorization prompt that maps diseases to one of six clinical systems (or \texttt{Other}) and returns a JSON tuple \{\texttt{primary\_diagnosis}, \texttt{category}\}. This supports the balanced per-system distribution in Table~\ref{tab1}.}
\label{figA5}
\end{figure*}

\section{Interaction and evaluation prompts}\label{AppB}

\subsection{Opening utterances.}
To initiate the encounter, the clinician uses one of \textbf{15} templated openings (Fig.~\ref{figB1}); these variants reduce prompt overfitting to any particular phrasing.
Upon the first opening, the SP-sim must return \textbf{exactly} the \texttt{1.Patient Information} and \texttt{2.Chief Complaint} sections from the structured record produced in the previous stage, without any inference or additional details.

\begin{figure*}[t]
\centering
\begin{minipage}{\textwidth}
\begin{tcolorbox}[
  colback=gray!3,
  colframe=gray!60!black,
  title=\textbf{Opening Utterances (15 Templates)},
  fonttitle=\bfseries,
  coltitle=black,
  boxrule=0.6pt,
  arc=1mm,
  left=1.5mm,
  right=1.5mm,
  top=1.5mm,
  bottom=1.5mm,
  sharp corners=south,
]
\footnotesize
\begin{itemize}
\item ``Hello, what brings you in today?''
\item ``Can you tell me the reason for your visit?''
\item ``Let’s begin — have you noticed anything unusual lately?''
\item ``What symptoms have you been experiencing?''
\item ``What seems to be the problem today?''
\item ``What kind of health concerns are you having recently?''
\item ``Please describe any discomfort or issues you've been having.''
\item ``Is there anything in particular that's been bothering you?''
\item ``Tell me what's going on — when did it start?''
\item ``Are you here because of any new or ongoing symptoms?''
\item ``What made you decide to come see a doctor today?''
\item ``We’ll start with your concerns — what would you like to discuss?''
\item ``Have you experienced any recent changes in your health?''
\item ``Can you walk me through your current symptoms?''
\item ``What’s the main reason for your appointment today?''
\end{itemize}
\end{tcolorbox}
\end{minipage}
\caption{Fifteen interchangeable opening utterances used by the clinician to start the encounter. On the first turn, the SP-sim responds with \texttt{1.Patient Information} and \texttt{2.Chief Complaint} verbatim from the structured case.}
\label{figB1}
\end{figure*}

\subsection{Standardized patient simulator (SP-sim)}
The standardized patient simulator described in Section~4.2 is instantiated with the following system prompt:\ref{figB2}

\begin{figure*}[t]
\centering
\begin{minipage}{\textwidth}
\begin{tcolorbox}[
  colback=gray!3,
  colframe=gray!60!black,
  title=\textbf{SP-sim Policy Prompt (Patient Role)},
  fonttitle=\bfseries,
  coltitle=black,
  boxrule=0.6pt,
  arc=1mm,
  left=1.5mm,
  right=1.5mm,
  top=1.5mm,
  bottom=1.5mm,
  sharp corners=south,
]
\footnotesize
You are simulating a real patient during a clinical consultation. You will \textbf{ONLY} answer based on this record, and \textbf{ONLY} respond to the doctor’s most recent statement. \\

\textbf{----- Case Record -----} \\
\texttt{{formatted\_record}} \\

\textbf{Opening Behavior:} \\
When the consultation begins, the doctor will ask a general opening question (e.g., “What brings you in today?” or “Where do you feel unwell?”). \\
In your first response, you must present the following two sections \textbf{exactly as written in the case record}: \\
- \textbf{1.Patient Information}  \\
- \textbf{2.Chief Complaint} \\

\textbf{After the opening, respond ONLY when the doctor requests a specific module or test.} \\

\textbf{Allowed Module Responses:} \\
- [History of Present Illness] \\
- [Past Medical History] \\
- [Physical Examination] \\

\textbf{Allowed Examination Requests (you will recognize them like this):} \\
- [Laboratory Tests: ...] \\
- [Imaging Studies: ...] \\
- [Functional Tests: ...] \\
- [Specialized Panels: ...] \\

\textbf{How to respond:}
\begin{itemize}
  \item Only respond if the requested test \textbf{is present in the case record}.
  \item Return the exact predefined result from the record.
  \item If a test is \textbf{not included in the case}, reply with: \\
  \texttt{"This test was not performed yet."}
  \item \textbf{Do NOT} invent any additional information.
  \item \textbf{Do NOT} include or hint at the final diagnosis at any time.
\end{itemize}

\textbf{Response Format:} \\
\texttt{[Module Name]: Full original text from that section} \\
\texttt{[test\_name]: Full original test\_result from that section}
\end{tcolorbox}
\caption{Policy prompt for the standardised patient driven by Qwen2.5-32B-Instruct. The simulator is reactive, releases only requested modules/tests, echoes verbatim from the record, and never reveals the diagnosis.}
\label{figB2}
\end{minipage}
\end{figure*}

\subsection{Doctor-agent configuration and action templates}
We run \textbf{15} instruction-tuned LLMs as the doctor agent . 
To ensure strict comparability, all models share deterministic decoding and identical prompt scaffolds (Figs.~\ref{figB3} \ref{figB4}). 

\begin{figure*}[t]
\centering
\begin{minipage}{\textwidth}
\begin{tcolorbox}[
  colback=gray!3,
  colframe=gray!60!black,
  title=\textbf{Task~1 — Full-Context Clinician Prompt},
  fonttitle=\bfseries,
  coltitle=black,
  boxrule=0.6pt,
  arc=1mm,
  left=1.5mm,
  right=1.5mm,
  top=1.5mm,
  bottom=1.5mm,
  sharp corners=south,
]
\footnotesize
You are an experienced senior clinician at a top-tier tertiary hospital. Your task is to carefully analyze structured patient records and provide the most accurate final diagnosis based on the clinical information. \\

You must follow strict clinical reasoning and adhere to the output format and diagnostic criteria described below. \\

\textbf{Output Format} \\
- When reaching a final diagnosis, \textbf{YOU MUST} start the response with: \\
\texttt{[Final Diagnosis] [Diagnosis Name]. Confirmed by:} \\
\texttt{1. ...} \\
\texttt{2. ...} \\
\texttt{3. ...} \\

- You \textbf{MUST} include the exact tag \texttt{[Final Diagnosis]} with brackets — do not rephrase, omit, or replace it.
\end{tcolorbox}
\caption{Clinician prompt for \textbf{Task~1} (full context). The agent reads the complete structured record and must output \texttt{[Final Diagnosis]} followed by exactly three evidential items, using the mandatory tag verbatim.}
\label{figB3}
\end{minipage}
\end{figure*}

\begin{figure*}[t]
\centering
\begin{minipage}{\textwidth}
\begin{tcolorbox}[
  colback=gray!3,
  colframe=gray!60!black,
  title=\textbf{Task~2 — Active Evidence-Seeking Clinician Prompt},
  fonttitle=\bfseries,
  coltitle=black,
  boxrule=0.6pt,
  arc=1mm,
  left=1.5mm,
  right=1.5mm,
  top=1.5mm,
  bottom=1.5mm,
  sharp corners=south,
]
\footnotesize
You are an experienced senior clinician at a top-tier tertiary hospital. Your goal is to gather only the necessary clinical information to reach a diagnosis. \\

You may choose \textbf{ONE} action per turn from:
\begin{itemize}
  \item \texttt{[History of Present Illness]}
  \item \texttt{[Past Medical History]}
  \item \texttt{[Physical Examination]}
  \item Request \texttt{[Laboratory Tests: test\_name]}
  \item Request \texttt{[Imaging Studies: test\_name]}
  \item Request \texttt{[Functional Tests: test\_name]}
  \item Request \texttt{[Specialized Panels: test\_name]}
  \item \texttt{[Final Diagnosis]}
\end{itemize}

\textbf{Your output format should be:} \\
\texttt{(one of the actions): (your reasoning process for choosing the next action)} \\

\textbf{Output Format for Final Diagnosis:}
\begin{itemize}
  \item When reaching a final diagnosis, \textbf{YOU MUST} start the response with: \\
  \texttt{[Final Diagnosis] [Diagnosis Name]. Confirmed by:} \\
  \texttt{1. ...} \\
  \texttt{2. ...} \\
  \texttt{3. ...}
  \item You \textbf{MUST} include the exact tag \texttt{[Final Diagnosis]} with brackets — do not rephrase, omit, or replace it.
\end{itemize}

\textbf{Note:}
\begin{itemize}
  \item To improve diagnostic efficiency, please perform tests only when necessary for diagnosis.
  \item You may only request \textbf{one specific test per turn}.
  \item \textbf{Do NOT} repeat tests or other modules.
  \item When confident, issue a \texttt{[Final Diagnosis]}.
  \item You must complete the diagnosis within a maximum of \textbf{10 turns}.
  \item If a test or module returns \texttt{"This test was not performed yet."}, you \textbf{MUST NOT} request it again.
  \item If your previous request failed (e.g., returned \texttt{"This test was not performed yet."}), consider alternatives or proceed to diagnosis if sufficient.
  \item For each turn, check previous requests before proposing the next action.
  \item \textbf{Do not re-request} the same test, even if it previously failed.
\end{itemize}
\end{tcolorbox}
\caption{Clinician prompt for \textbf{Task~2} (active evidence-seeking). Each turn, the agent selects exactly one action (module or a specific test), avoids repetitions and unavailable tests, and stops with \texttt{[Final Diagnosis]} within 10 turns. For non-terminal turns it outputs \texttt{(action): (brief rationale)}; final outputs must include the mandatory tag and three supporting items.}
\label{figB4}
\end{minipage}
\end{figure*}

\paragraph{Endpoints and scoring.}
We report two endpoints: \emph{Final Diagnosis Accuracy} and \emph{Evidence Quality}.
Accuracy is scored on a 3-level scale \{0,1,2\}: \textbf{0} incorrect (different disease family), \textbf{1} partially correct (right family, wrong subtype/term), \textbf{2} fully correct (including synonyms/subtypes).
Evidence Quality is also scored on \{0,1,2\}: \textbf{2} all three evidence items are supported and consistent; \textbf{1} one–two items supported or partially reasonable; \textbf{0} unsupported/contradictory.
Figures~\ref{figB5} and~\ref{figB6} show the prompts used to elicit \emph{single-integer} decisions.

\begin{figure*}[t]
\centering
\begin{minipage}{\textwidth}
\begin{tcolorbox}[
  colback=gray!3,
  colframe=gray!60!black,
  title=\textbf{Accuracy Evaluation Prompt (0/1/2)},
  fonttitle=\bfseries,
  coltitle=black,
  boxrule=0.6pt,
  arc=1mm,
  left=1.5mm,
  right=1.5mm,
  top=1.5mm,
  bottom=1.5mm,
  sharp corners=south,
]
\footnotesize
You are a senior medical expert tasked with evaluating whether the AI model's diagnosis is accurate. Score according to the criteria below:

\textbf{[Scoring Criteria]}
\begin{itemize}
  \item \textbf{0}: Completely incorrect (a different disease category)
  \item \textbf{1}: Partially correct (correct disease category, but wrong subtype)
  \item \textbf{2}: Completely correct (including synonyms or correct subtype)
\end{itemize}

\textbf{[Output Requirements]} \\
Return only a single integer score (0, 1, or 2). \textbf{Do NOT} include any other text.
\end{tcolorbox}
\caption{Prompt used by the external evaluator to score \textbf{Final Diagnosis Accuracy} on a 0/1/2 scale.
The evaluator returns a \emph{single integer} only.
This prompt is used identically for \textbf{Task~1} and \textbf{Task~2}.}
\label{figB5}
\end{minipage}
\end{figure*}

\begin{figure*}[t]
\centering
\begin{minipage}{\textwidth}
\begin{tcolorbox}[
  colback=gray!3,
  colframe=gray!60!black,
  title=\textbf{Evidence Quality Evaluation Prompt (0/1/2)},
  fonttitle=\bfseries,
  coltitle=black,
  boxrule=0.6pt,
  arc=1mm,
  left=1.5mm,
  right=1.5mm,
  top=1.5mm,
  bottom=1.5mm,
  sharp corners=south,
]
\footnotesize
You are a medical evidence expert tasked with evaluating how well the diagnostic evidence provided by the model matches the clinical case information. Please assign a score based on the following criteria:

\textbf{[Scoring Criteria]}
\begin{itemize}
  \item \textbf{2}: All 3 evidence points are clearly supported in the case and fully consistent with the diagnosis
  \item \textbf{1}: 1–2 evidence points are supported, or there is partial reasonable inference
  \item \textbf{0}: Evidence points contradict the case or are not supported at all
\end{itemize}

\textbf{[Output Requirement]} \\
Only return a single integer score (0, 1, or 2). \textbf{DO NOT} include any other text.

\textbf{[Important]}
\begin{itemize}
  \item Only evaluate the model's evidence points (usually 1–3)
  \item Evidence must be specific, verifiable, and clearly stated
\end{itemize}
\end{tcolorbox}
\caption{Prompt used by the external evaluator to score \textbf{Evidence Quality} on a 0/1/2 scale.
The evaluator checks that the model's three evidence items are \emph{verbatim supported} by the case and consistent with the diagnosis, and returns a \emph{single integer} only.
This prompt is used identically for \textbf{Task~1} and \textbf{Task~2}.}
\label{figB6}
\end{minipage}
\end{figure*}




\end{appendices}



\end{document}